\PassOptionsToPackage{table}{xcolor}

\documentclass[sigconf]{acmart}
\AtBeginDocument{%
  }
\usepackage{tcolorbox}

\usepackage{graphicx}

\usepackage{multirow}
\usepackage{pifont}
\usepackage{colortbl}
\usepackage{booktabs}
\usepackage[ruled,linesnumbered]{algorithm2e}

\usepackage{soul}
\usepackage{fontawesome} 
\usepackage{graphicx}
\usepackage{enumitem}
\usepackage{dsfont}

\copyrightyear{2025}
\acmYear{2025}
\setcopyright{acmlicensed}\acmConference[MM '25]{Proceedings of the 33rd ACM International Conference on Multimedia}{October 27--31, 2025}{Dublin, Ireland}
\acmBooktitle{Proceedings of the 33rd ACM International Conference on Multimedia (MM '25), October 27--31, 2025, Dublin, Ireland}
\acmDOI{10.1145/3746027.3755064}
\acmISBN{979-8-4007-2035-2/2025/10}

\begin{document}

\title{Uni-Layout: Integrating Human Feedback in Unified Layout Generation and Evaluation}

\author{Shuo Lu}
\orcid{0009-0000-7547-3169}
\affiliation{%
  \institution{NLPR \& MAIS, CASIA; School of AI, UCAS}
  \city{Beijing}
  \country{China}
}
\email{shuolucs@gmail.com}
\authornote{Work done while interning at JD.COM}

\author{Yanyin Chen}
\orcid{0009-0004-5508-3605}
\affiliation{
  \institution{JD.COM}
  \city{Beijing}
  \country{China}
}
\email{chenyanyin6@jd.com}
\authornote{The first two authors contributed equally to this research.}

\author{Wei Feng}
\orcid{0009-0005-8890-4956}
\affiliation{%
  \institution{JD.COM}
  \city{Beijing}
  \country{China}
}
\email{fengwei25@jd.com}

\author{Jiahao Fan}
\orcid{0009-0004-4665-7152}
\affiliation{%
  \institution{JD.COM}
  \city{Beijing}
  \country{China}
}
\email{fanjiahao5@jd.com}

\author{Fengheng Li}
\orcid{0009-0004-2145-9256}
\affiliation{%
  \institution{JD.COM}
  \city{Beijing}
  \country{China}
}
\email{lifengheng6@jd.com}

\author{Zheng Zhang}
\orcid{0009-0002-6391-4814}
\affiliation{%
  \institution{JD.COM}
  \city{Beijing}
  \country{China}
}
\email{zhangzheng11@jd.com}

\author{Jingjing Lv}
\orcid{0009-0000-5518-7077}
\affiliation{%
  \institution{JD.COM}
  \city{Beijing}
  \country{China}
}
\email{lvjingjing1@jd.com}

\author{Junjie Shen}
\orcid{0009-0008-6983-5213}
\affiliation{%
  \institution{JD.COM}
  \city{Beijing}
  \country{China}
}
\email{shenjunjie@jd.com}

\author{Ching Law}
\orcid{0009-0001-3275-2528}
\affiliation{%
  \institution{JD.COM}
  \city{Beijing}
  \country{China}
}
\email{lawching@jd.com}

\author{Jian Liang}
\orcid{0000-0003-3890-1894}
\affiliation{%
  \institution{NLPR \& MAIS, CASIA; School of AI, UCAS}
  \city{Beijing}
  \country{China}
}
\email{liangjian92@gmail.com}
\authornote{Corresponding author.}

\renewcommand{\shortauthors}{Lu et al.}

\begin{abstract}
Layout generation plays a crucial role in enhancing both user experience and design efficiency. 
However, current approaches suffer from task-specific generation capabilities and perceptually misaligned evaluation metrics, leading to limited applicability and ineffective measurement.
In this paper, we propose \textit{Uni-Layout}, a novel framework that achieves unified generation, human-mimicking evaluation and alignment between the two.
For universal generation, we incorporate various layout tasks into a single taxonomy and develop a unified generator that handles background or element contents constrained tasks via natural language prompts.
To introduce human feedback for the effective evaluation of layouts, we build \textit{Layout-HF100k}, the first large-scale human feedback dataset with 100,000 expertly annotated layouts.
Based on \textit{Layout-HF100k}, we introduce a human-mimicking evaluator that integrates visual and geometric information, employing a Chain-of-Thought mechanism to conduct qualitative assessments alongside a confidence estimation module to yield quantitative measurements.
For better alignment between the generator and the evaluator, we integrate them into a cohesive system by adopting Dynamic-Margin Preference Optimization (DMPO), which dynamically adjusts margins based on preference strength to better align with human judgments.
Extensive experiments show that \textit{Uni-Layout} significantly outperforms both task-specific and general-purpose methods.
Our code is publicly available at \url{https://github.com/JD-GenX/Uni-Layout}.
\end{abstract}

\ccsdesc[500]{Computing methodologies~Computer vision}


\keywords{Unified Layout Generation and Evaluation, Reinforcement Learning from Human Feedback, Multimodal Large Language Models}


\maketitle

\section{Introduction}

\begin{figure*}
    \centering
    \includegraphics[width=0.95\textwidth]{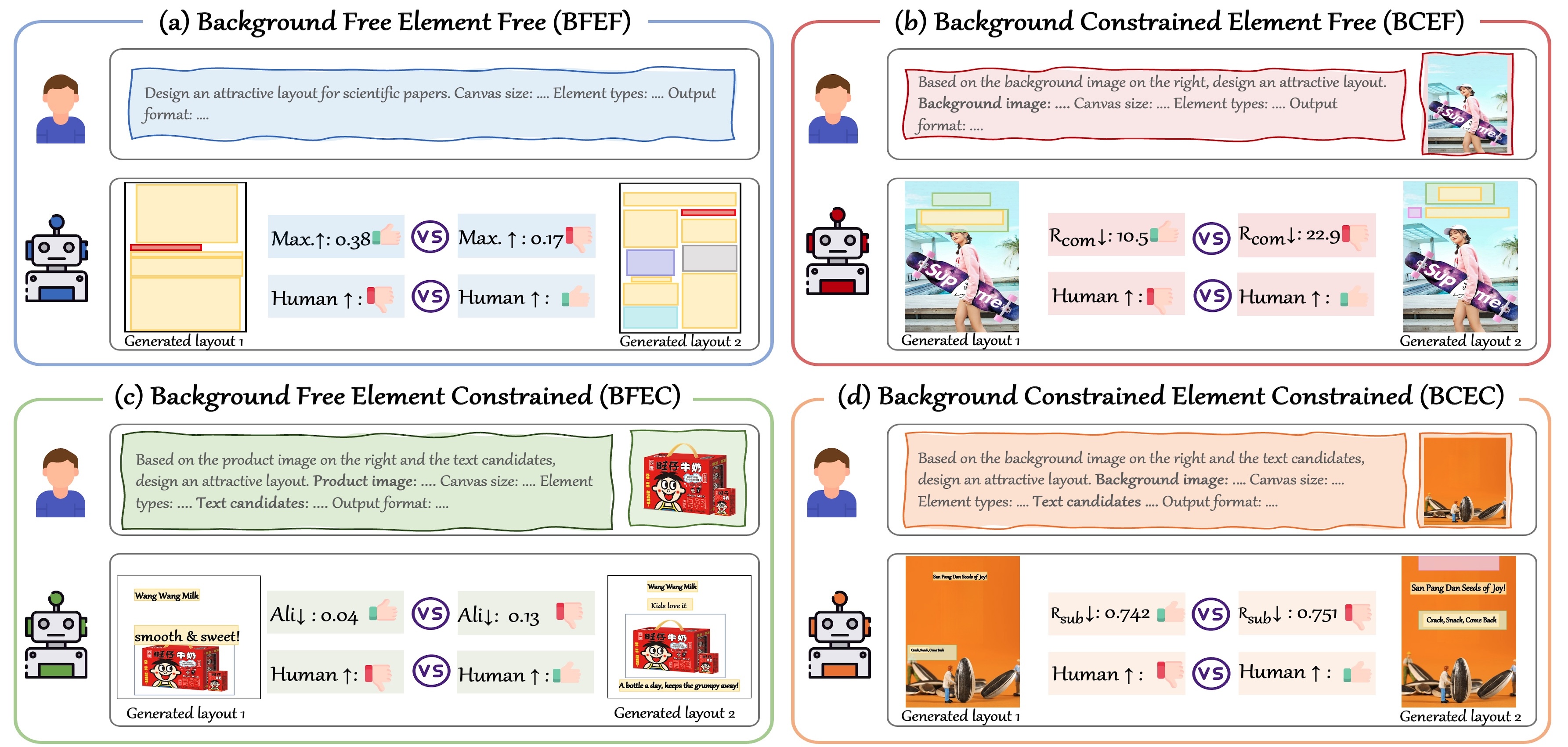}
    \caption{Taxonomy of layout generation tasks and illustration of motivation. Diverse layout generation tasks can be divided into four categories: (a) BFEF, (b) BCEF, (c) BFEC and (d) BCEC. Different tasks require different models (\protect\includegraphics[height=1.2em]{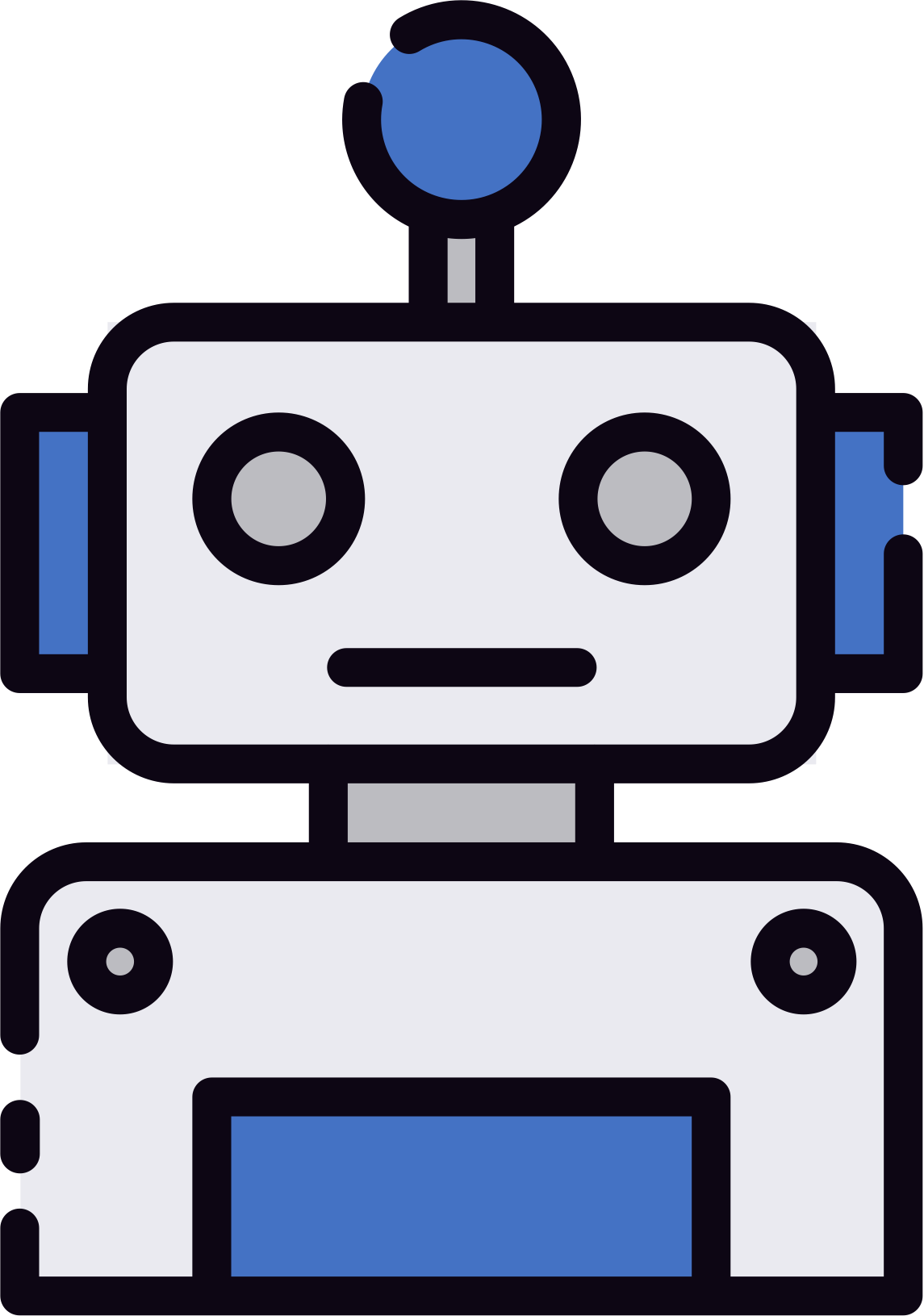}/\protect\includegraphics[height=1.2em]{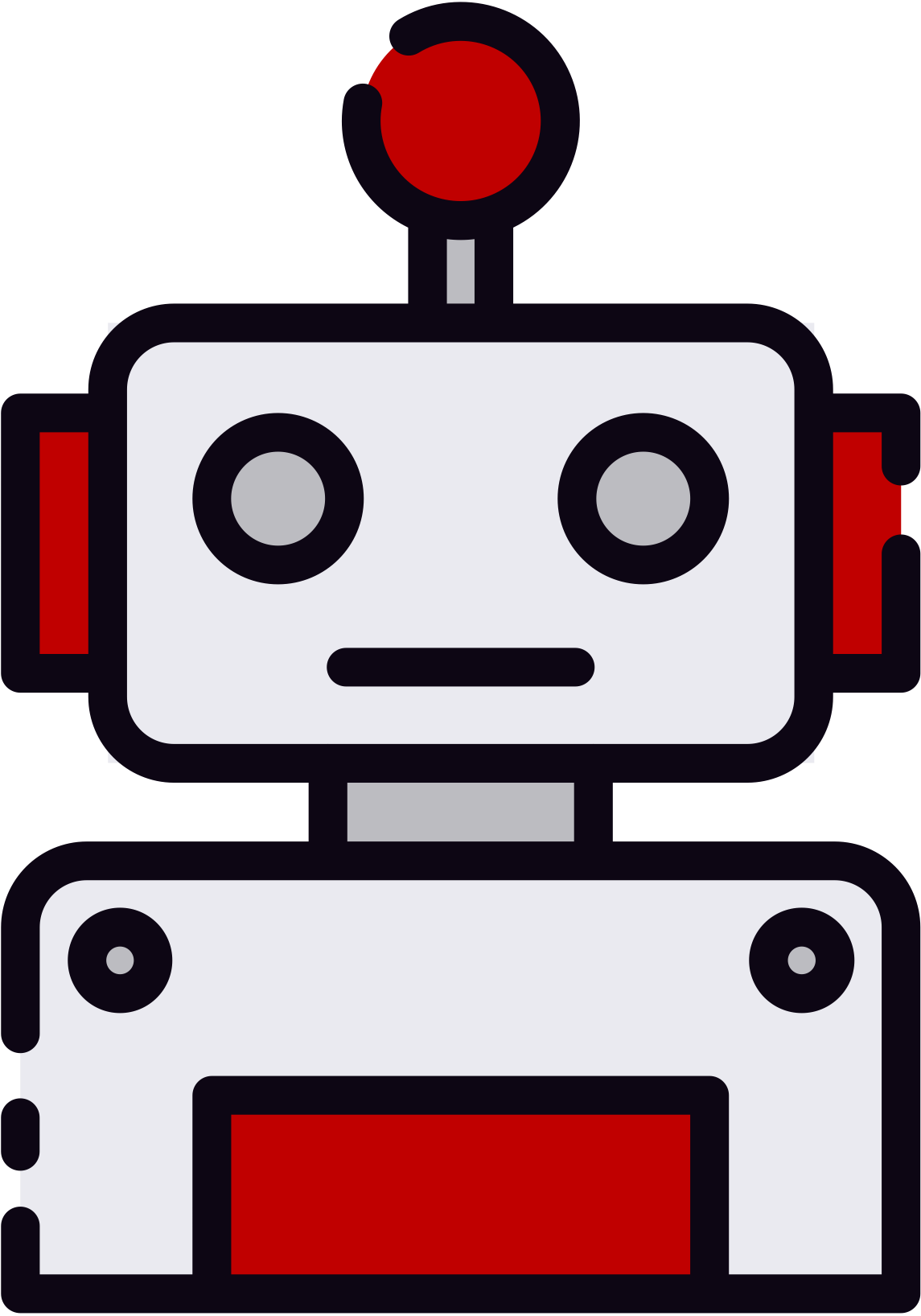}/\protect\includegraphics[height=1.2em]{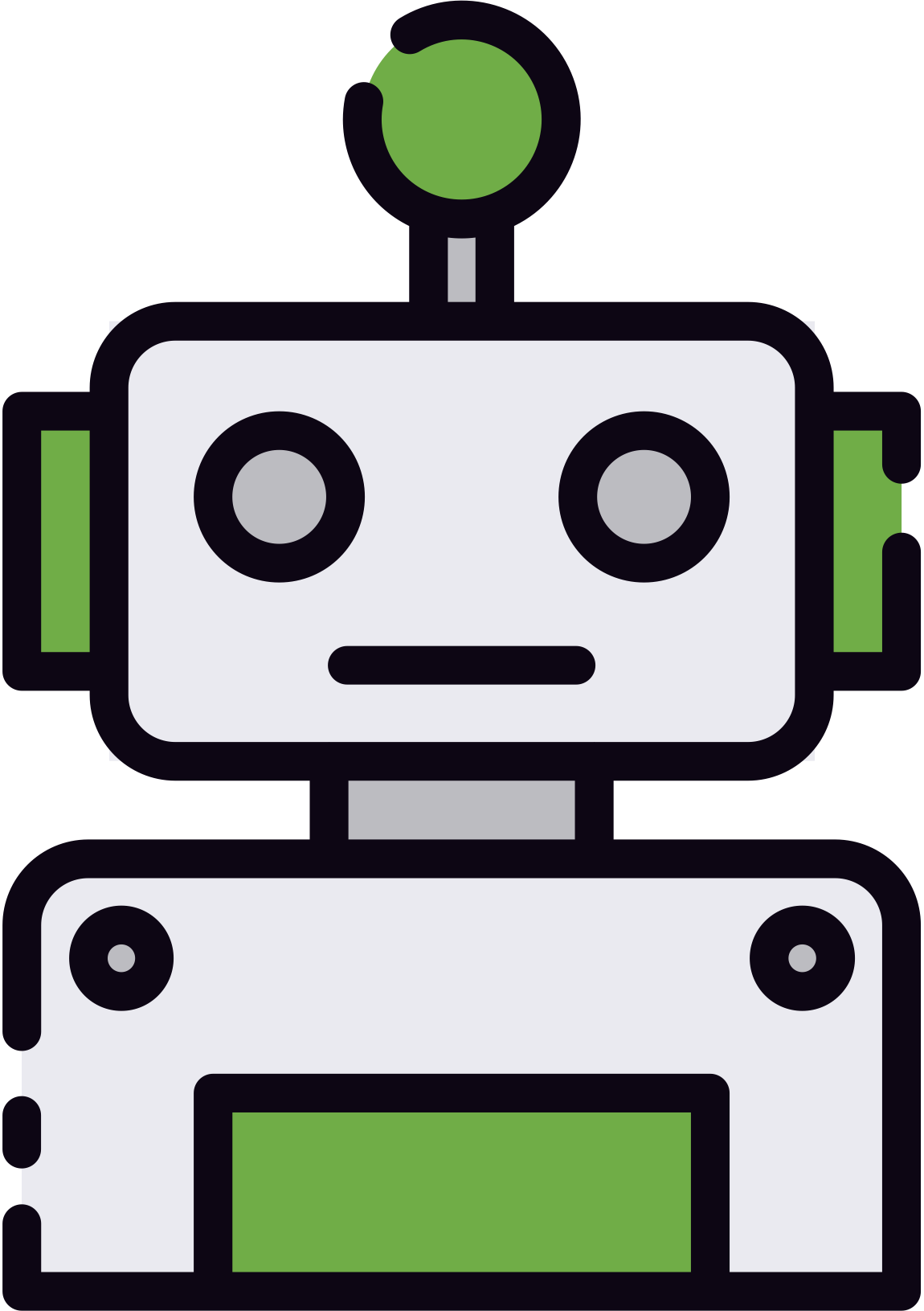}/\protect\includegraphics[height=1.2em]{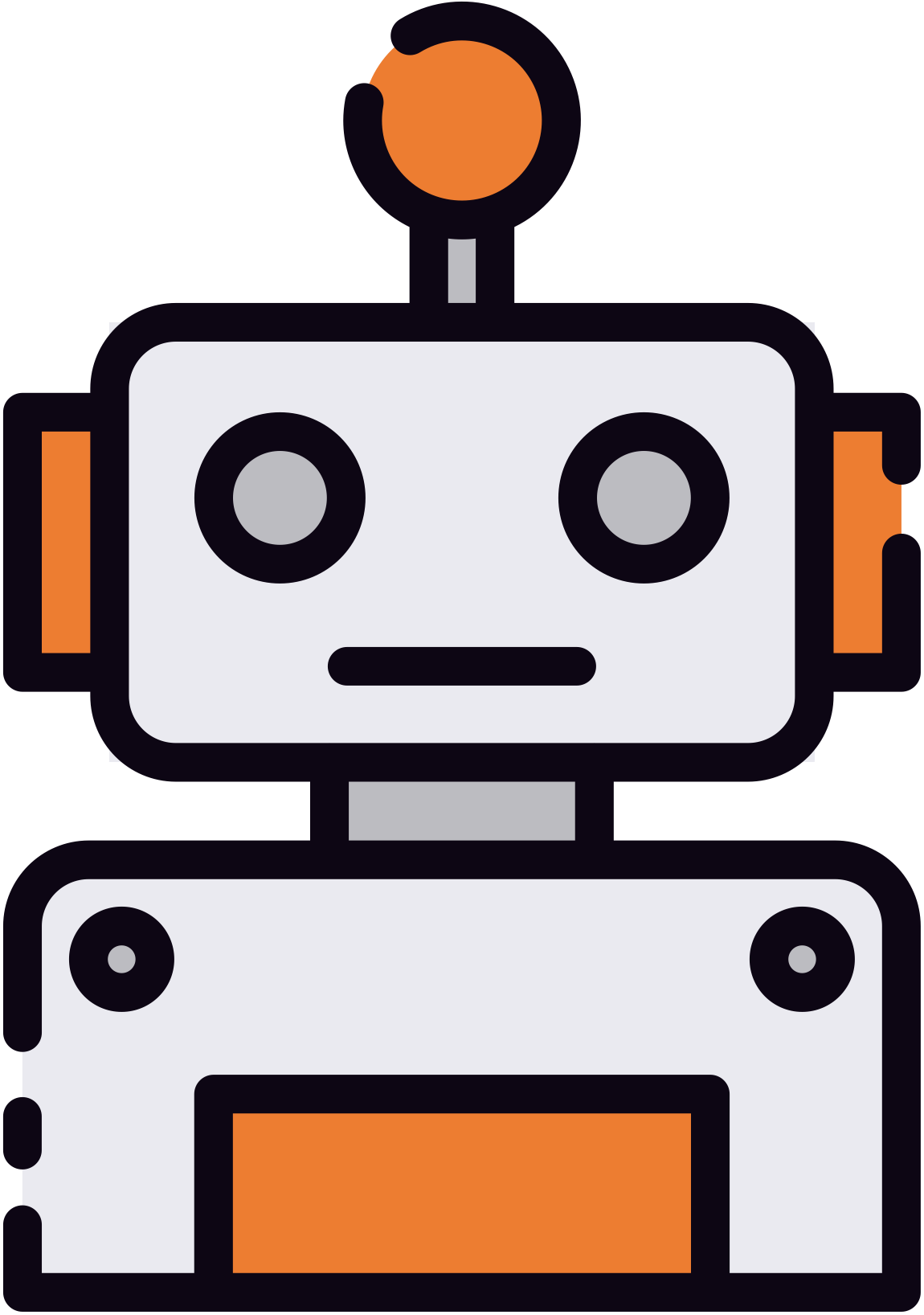}), which lacks flexibility. Within each task, there exists a discrepancy between metric performance and human perception: (``Generated layout 1'') better metric but poor human-perceived layouts vs. (``Generated layout 2'') worse metric but human-preferred layouts.}
    \label{fig:fig1}
\end{figure*}

Layout generation, the process of creating visually appealing and practical arrangements based on given information~\citep{lok2001survey}, has emerged as a fundamental component in modern design systems. 
It covers a broad spectrum of tasks, ranging from poster and document design \citep{hsu2023posterlayout,he2023diffusion} to user interface layouts and magazine composition\citep{raneburger2012automated,tabata2019automatic}.

Despite significant progress, existing layout generation methods typically focus on narrowly-defined tasks~\citep{li2023relation, li2024design, zhang2024creposter, li2023planning,zhou2022composition,li2020attribute}, leading to task-specific solutions that lack flexibility and generalizability. 
Worse still, although existing evaluation metrics are carefully designed upon meaningful layout design principles (e.g., reducing visual clutter~\citep{zhou2022composition} or enhancing alignment~\citep{arroyo2021variational}), they often misalign with human perception.
As illustrated in Figure~\ref{fig:fig1}, 
layouts with  higher scores may exhibit inferior visual qualities, revealing the gap between existing metrics and genuine human perception.
To address these challenges, we introduce \textit{Uni-Layout}, a holistic framework that unifies layout generation, evaluation, and alignment through three interconnected components: a unified generator, a unified human-mimicking evaluator, and a dynamic-margin alignment mechanism. 
To elaborate on \textit{Uni-Layout}, we structure our framework around three core research questions.


\textbf{\normalsize \faQuestionCircle \ How can we achieve unified layout generation across diverse tasks?}
To systematically unify the currently fragmented landscape of layout generation tasks~\citep{li2024design,zhang2024creposter,hsu2023posterlayout,weng2024desigen,chen2024iris,xu2024image,wang2024prompt2poster,zhong2025scientific,guo2025contentdm}, we propose a well-organized taxonomy based on two dimensions: whether the \textbf{B}ackground and \textbf{E}lement contents  are \textbf{F}ree or \textbf{C}onstrained.
As shown in Figure ~\ref{fig:fig1}, 
we group existing layout tasks into four representative types~\footnote{The four tasks align with the poster creation workflow. For instance, BFEC integrates with background generation for poster creation, while BCEF can be combined with text image generation. Layout generation underpins aesthetic quality and visual harmony.}: 
\textit{BFEF}~\citep{chen2024towards,arroyo2021variational,inoue2023layoutdm,kong2022blt,tang2023layoutnuwa,guerreiro2024layoutflow,kikuchi2024multimodal,li2020attribute}, \textit{BCEF}~\citep{horita2024retrieval,hsu2023posterlayout,zhou2022composition,seol2024posterllama,guo2025contentdm},  \textit{BFEC}~\citep{li2023planning,chen2025paid,zhang2024creposter} and  \textit{BCEC}~\citep{li2023relation,li2024design,seol2024posterllama}. 
Current task-specific methods struggle with unified layout generation, but multimodal large language models~\citep{liu2023visual} (MLLMs) offer promising solutions due to their general visual-language understanding capabilities.
Leveraging MLLMs, we propose a unified layout generator that works like a skilled designer. It integrates visual constraints and textual instructions to generate coherent layouts, handling diverse scenarios where both background and element contents can be either constrained or free.
Benefiting from joint training across various layout tasks, it provides a flexible and unified solution for layout generation.

\textbf{\normalsize \faQuestionCircle \ How can we effectively evaluate layouts following human judgment patterns?} 
In spite of the importance of human perception in layout design, human feedback on layout quality is absent in existing datasets~\citep{zhou2022composition,li2023relation,zhong2019publaynet}.
To bridge this gap, we aggregate outputs from the unified generator and compile \textit{Layout-HF100k}, the first-of-its-kind comprehensive human feedback dataset curated for layout generation, featuring 100k meticulously annotated high-quality examples across representative layout tasks.
Based on the brand-new dataset, we develop an evaluator that processes layouts through dual branches - visual and geometric information - to effectively mimic human judgment patterns.
Furthermore, the evaluator incorporates a classification head that outputs quantitative confidence estimations, alongside qualitative Chain-of-Thought (CoT) reasoning~\citep{wei2022chain}, enabling it to capture subtle aesthetic preferences and provide interpretable assessments that closely align with human perceptual patterns.
By combining multi-modal analysis with CoT reasoning, our evaluator not only makes accurate judgments but also articulates the rationale behind its decisions, similar to how human experts evaluate layouts.

\textbf{\normalsize \faQuestionCircle \ How can we effectively align layout generation with human feedback?}
Existing alignment methods either directly maximize the likelihood of human-preferred outputs ~\citep{rafailov2023direct, song2024preference}, or employ fixed margins ~\citep{zeng2024token, meng2024simpo} in their preference learning objectives. 
Such conventional methods fail to reflect the varying degrees of human preferences, as they treat strong and weak preferences equally.
To address this limitation, we propose a novel alignment method called Dynamic-Margin Preference Optimization (\textit{DMPO}).
Concretely, when evaluators express stronger preferences between paired samples, \textit{DMPO} automatically increases the margin to enforce larger score differences between winning and losing responses, while applying smaller margins for less distinct preferences. This confidence-guided adaptive margin strategy better captures the spectrum of human judgments, leading to more precise alignment with layout generation and human preferences.

In summary, our contributions are threefold:

\begin{itemize}
\item We introduce \textit{Uni-Layout}, a unified framework that systematically integrates layout generation, evaluation, and alignment. Under this framework, we propose an instruction-driven unified layout generator capable of effectively handling diverse layout tasks.

\item We present \textit{Layout-HF100k}, the first comprehensive human preference dataset for layout generation, containing 100k high-quality instances. 
Based on \textit{Layout-HF100k}, we develop a human-mimicking evaluation with dual-branch evaluation mechanism enhanced by qualitative CoT reasoning, closely aligning with human judgments.

\item We propose \textit{DMPO}, a dynamic-margin alignment method that tightly couples layout generation and evaluation, achieving better alignment with human preferences and significantly outperforming existing baseline methods in extensive experiments.
\end{itemize}

\begin{figure*}
    \centering
    \includegraphics[width=0.90\linewidth]{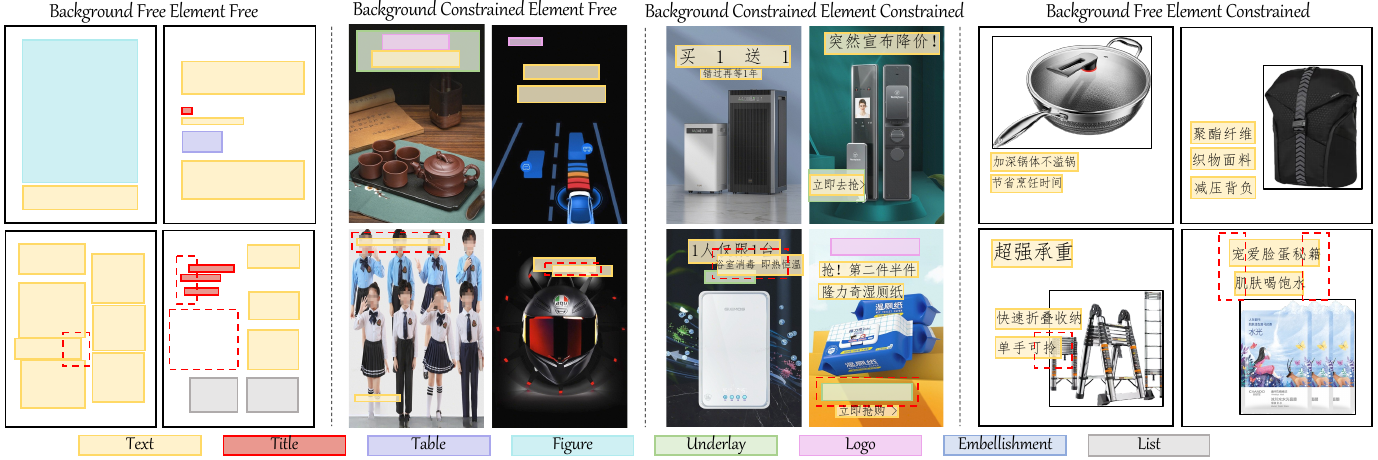}
    \caption{Layout-HF100k examples. The top row shows qualified examples, while the bottom row shows unqualified ones. The unqualified parts are noted by the red dashed line (the same below).}
    \label{fig:dataset}
\end{figure*}

\section{Related Work}
\subsection{ Layout Generation}

With the advancement of content generation technologies~\citep{tang2023layoutnuwa, zhang2024artbank, jing2023vision}, researchers continue to explore novel techniques to address diverse layout scenarios, typically focusing on outputting the positions and types of bounding boxes. Existing methods can be categorized into four classes depending on whether the background and element contents are constrained.

\noindent
\textbf{Background-Free and Element-Free (BFEF)} methods \cite{kong2022blt, inoue2023layoutdm, chen2024towards, guerreiro2024layoutflow} are widely used in document and web design. 
Recent diffusion models \cite{jiang2023res,feng2024fancyvideo,wang2024qihoo,cao2025relactrl,wang2025wisa} and flow matching \cite{guerreiro2024layoutflow} handle structured layout data effectively. 
Evaluation metrics for BFEF tasks typically include geometric measurements like maximum Intersection-over-Union (Max.)~\citep{kikuchi2021constrained}, alignment~\citep{li2020attribute}, and overlap~\citep{hsu2023posterlayout}.

\noindent
\textbf{Background-Constrained and Element-Free (BCEF)} methods \cite{guo2025contentdm, seol2024posterllama, hsu2023posterlayout, zhou2022composition} are mainly used in poster layout design, focusing on background comprehension. Some GAN-based methods \cite{zhou2022composition, hsu2023posterlayout} employ encoder-decoder architectures for visual feature extraction but lack semantic integration. 
Current BCEF metrics extend beyond geometric measurements to include background coordination and occlusion, such as $R_{com}$, $R_{sub}$, and $R_{occ}$~\citep{zhou2022composition, li2023relation,he2025plangen}.

\noindent
\textbf{Background-Free and Element-Constrained (BFEC)} methods \cite{chen2025paid, li2023planning, zhang2024creposter} are prominent in product poster generation, focusing on element content comprehension. Unlike BFEF, BFEC emphasizes understanding element contents. P\&R \cite{li2023planning} uses a diffusion model framework with ViT-B/16 \cite{dosovitskiy2020image} and RoBERTa \cite{liu2019roberta} for encoding. 
BFEC methods are evaluated with metrics similar to BFEF.

\noindent
\textbf{Background-Constrained and Element-Constrained (BCEC)} methods \cite{seol2024posterllama, li2024design, li2023relation} are used in poster design, requiring simultaneous processing of background and layout elements. RADM \cite{li2023relation} learns cross-modal relationships and uses a Geometric Relation Awareness Module to model spatial dependencies. 
BCEC methods use evaluation metrics similar to BCEF.

Despite the efficacy of existing layout models and evaluation metrics, they lack generalizability and flexibility. To address these, we introduce a method with a unified layout generator for cross-task adaptability and standardized quality assessment via human feedback integration.

\subsection{Learning from Human Feedback}
Reinforcement Learning with Human Feedback (RLHF) uses human evaluations as reward signals to optimize models, proving effective for training large-scale language models \cite{achiam2023gpt, mei2024realhf, kulkarni11, ouyang2022training,chen2025ctr}. Since 2017, RLHF~\cite{christiano2017deep} has advanced with methods like Direct Preference Optimization (DPO) \cite{rafailov2023direct} and game theory integration \cite{munos2023nash}. InstructGPT \cite{ouyang2022training} fine-tunes GPT-3 using RLHF with human-ranked outputs. Recent developments include critique-based reward models and dynamic reward scaling \cite{zhang2025mm}. RLHF has also expanded to image generation, focusing on human feedback reward models \cite{xu2023imagereward, kirstain2023pick, wu2023human, xu2024visionreward, wang2025generate} and RLHF-based optimization for generative models \cite{black2023training, wallace2024diffusion, du2024towards}.

Despite advances, layout generation is still unexplored in this context. To address this, we curate Layout-HF100k, the first large-scale human feedback dataset for layout generation. Using this dataset, we develop a novel reinforcement learning method that validates human feedback efficacy in layout tasks.

\section{Layout-HF100k}\label{sec:datasets}

\subsection{Composition}
To collect human feedback on layouts, we trained a unified  generator to produce layouts.
For each task, we selected a representative dataset:
CGL-Dataset~\citep{zhou2022composition} for BCEF, CGL-Dataset V2~\citep{li2023relation} for BCEC, and PubLayNet~\citep{zhong2019publaynet} for BFEF.
Since no existing BFEC methods open-source their datasets, we collected a new dataset named E-commerce Poster Layout (EP-Layout), which will be released to facilitate progress in BFEC tasks.
Next, we briefly introduce the four datasets.

\noindent
\textbf{CGL-Dataset}~\citep{zhou2022composition} supports poster layout generation by determining element positions on a given background. It includes 60,548 advertising posters with annotated layouts for training and 1,000 for testing, collected from Taobao. Elements are categorized as logo, text, underlay, and embellishment.

\noindent
\textbf{CGL-Dataset V2}~\citep{li2023relation} is an extension of CGL-Dataset~\citep{zhou2022composition}. This updated version introduces an additional 1,035 testing samples. Beyond the basic category and coordinates, CGL-Dataset V2 incorporates the textual content, which refers to the Chinese selling point in advertising posters. 

\noindent
\textbf{PubLayNet} ~\citep{zhong2019publaynet} is aimed at document layout generation with no constraints on background or element content. It includes over 360,000 document images annotated. The dataset is divided into 340,391 for training, 11,858 for validation, and 11,983 for testing. Element types include text, title, list, table, and figure.

\noindent
\textbf{EP-Layout} comprises high-quality posters from a world-renowned e-commerce platform, focusing on positioning product images and text on a blank background. It includes 100,000 training samples and 1,000 testing samples, with elements like product masks and selling points. Covering common product categories, it is useful for tasks like layout-to-image generation and image understanding.

Based on our unified generator's output, professional annotators evaluate layout quality, creating our \textit{Layout-HF100k} dataset. It comprises 96,000 training and 4,000 testing samples. Training samples are distributed as follows: 31,000 for BCEF, 19,000 for BFEF, 19,000 for BFEC, and 27,000 for BCEC. 
Each task includes 1,000 test samples with a roughly 1:1 positive to negative ratio.

\subsection{Annotation}
As shown in Figure~\ref{fig:dataset}, labels are divided into two classes: qualified and unqualified layouts. Standards for these are based on task-specific characteristics. Qualified layouts feature proper alignment and organization, while unqualified ones often suffer from poor element arrangement and unreadable text. 

To ensure dataset quality, we implement a rigorous annotation workflow, with annotators having over 5 years of experience in graphic design. The process includes three stages: primary annotation, quality inspection, and statistical sampling audit. Initially, annotators categorize layouts as qualified or unqualified based on detailed guidelines and examples. During quality inspection, a separate team verifies all labels against the original data. In the final stage, at least 10\% of each batch is re-examined by senior auditors, ensuring a minimum accuracy rate of 98\%. 
Submissions below this threshold are returned for revision. Throughout, we identify common errors to update guidelines and retrain annotators.

\subsection{Characteristic}

\textbf{Large-Scale and New Domains}: Although human feedback datasets have gained attention recently, they remain limited in scale. For example, LLaVA-RLHF~\citep{sun2023aligning} contains only 10k samples, and RLHF-V~\citep{yu2024rlhf} has just 1.4k samples. Even machine-generated alignment datasets~\citep{li2023silkie,wang2024enhancing,zhang2024mavis} rarely exceed 100k samples. Additionally, human feedback in layout design is largely unexplored. To fill these gaps, we introduce Layout-HF100k, which pioneers large-scale human feedback collection specifically for layout design.

\noindent
\textbf{Diversity and Flexibility}: Layout-HF100k encompasses a broad range of layout types. This diversity enables models to address a variety of design tasks with different constraints, thereby enhancing their adaptability and practical utility.

\noindent
\textbf{Model Evaluation and Benchmarking}: Unlike conventional layout generation benchmarks that primarily rely on rule-based metrics, Layout-HF100k introduces a human-centered evaluation paradigm. This innovative benchmark aligns layouts with human preferences, which provides a more meaningful assessment of model performance.

\section{Method}
\begin{figure*}
    \centering
    \includegraphics[width=0.95\textwidth]{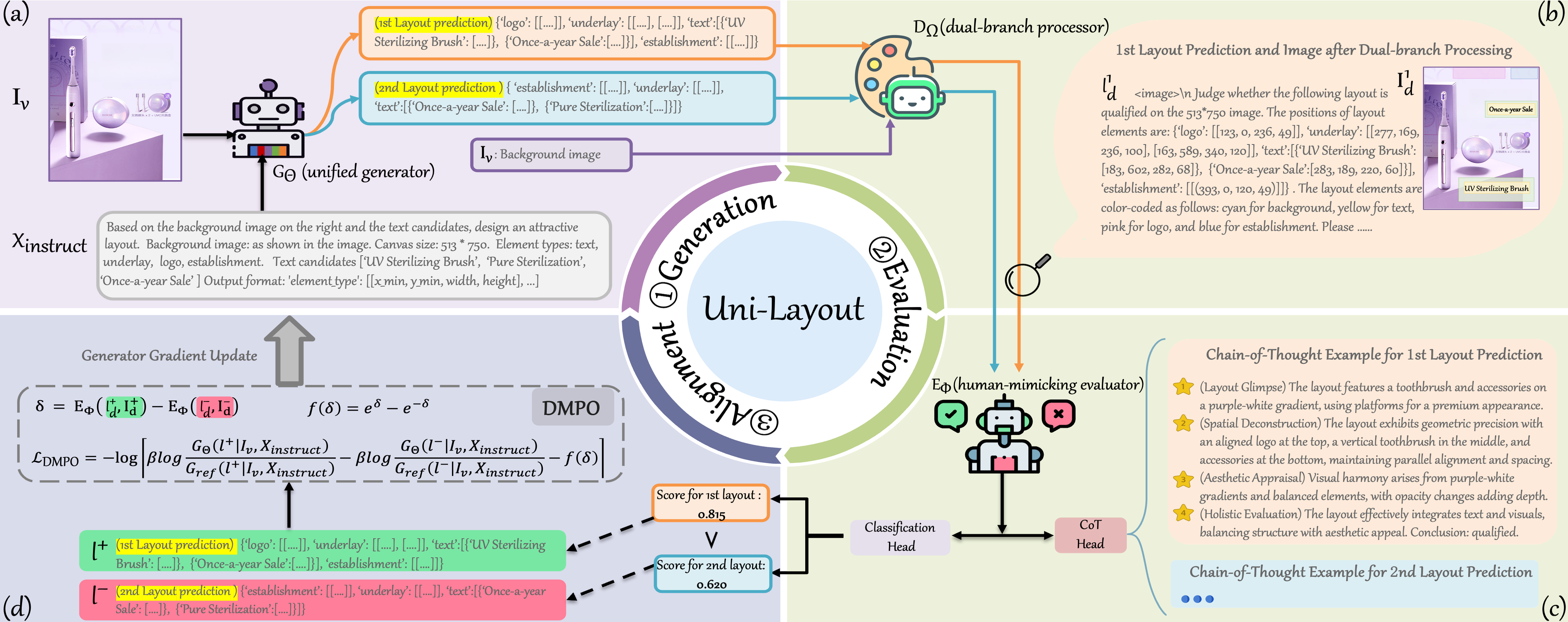}
    \caption{Overview of Uni-Layout framework: (a) Generation described in Section ~\ref{subsec:uni_generation}, (b,c) Evaluation in Section ~\ref{subsec:uni_evaluation}, and (d) Alignment in Section ~\ref{subsec:alignment}.  Here we use the Background-Constrained and Element-Constrained task as an example.}
    \label{fig:outline_overall}
\end{figure*}

\subsection{Overview}
In this work, we introduce \textit{Uni-Layout}, a comprehensive framework for layout generation, evaluation and alignment, as shown in Figure~\ref{fig:outline_overall}. We unify layout tasks into a taxonomy and pretrain a MLLM on a diverse dataset to create a unified generator. 
We then develop a unified layout evaluator that achieves human-like layout assessment capabilities through a dual-branch learning strategy and CoT reasoning, trained on the proposed \textit{Layout-HF100k} dataset.
Finally, we propose \textit{DMPO}  to align layout generation with human feedback, fine-tuning the generator based  on evaluator feedback to produce aesthetically pleasing layouts.

\subsection{Unified Generation}\label{subsec:uni_generation}

\subsubsection{Unified Instruction Design}
To systematically handle various layout tasks, we employ a unified multimodal instruction-following paradigm.
Given visual inputs (background or element images) denoted as $\mathbf{I}_\mathrm{v}$ and textual instructions $\mathbf{X}_\mathrm{instruct}$, the unified generator $G_{\Theta}$ produces layout predictions $l$ (an intuitive abbreviation for ``layout'') through a multimodal instruction-following process. Formally, the complete multimodal prompt is structured as follows:
\begin{equation}
\resizebox{0.9\columnwidth}{!}{
$\mathrm{\textit{Human}}:\mathbf{I}_\mathrm{v}<\backslash n\mathrm{>}~\mathbf{X}_\mathrm{instruct}\text{<STOP> \textit{Assistant}}:l\mathrm{<STOP>}$
}
\label{eq:multimodal_prompt}
\end{equation}
\noindent Here, \textit{Human} and \textit{Assistant} denote role tokens, while $<\backslash n>$ and $\mathrm{<STOP>}$ indicate line break and sequence termination, respectively~\footnote{For BFEF tasks without image inputs, omit $\mathbf{I}_\mathrm{v}$ and the $<\backslash n>$ tokens.}.

For the textual instruction part, $\mathbf{X}_\mathrm{instruct}$ is generated by our scalable instruction function $f_{\mathrm{instruct}}(\cdot)$, which systematically integrates various task-specific constraints into structured prompts. 
A layout task instruction is typically composed of six key components,  which can be formally defined as:
\begin{equation}
\mathbf{X}_\mathrm{instruct} = f_{\mathrm{instruct}}(\textcolor{green}{T}, \textcolor{blue}{b_a}, \textcolor{cyan}{b_c}, \textcolor{orange}{e_a^1}, \textcolor{orange}{e_a^2}, \ldots, \textcolor{orange}{e_a^m}, \textcolor{yellow}{e_c^1}, \textcolor{yellow}{e_c^2}, \ldots, \textcolor{yellow}{e_c^n}, \textcolor{red}{O}),
\label{eq:instruct}
\end{equation}
\noindent where $\textcolor{green}{T}$ is the task description, $\textcolor{blue}{b_a}$ represents background attributes, $\textcolor{cyan}{b_c}$ denotes background content, $\textcolor{orange}{e_a^1}, \textcolor{orange}{e_a^2}, \ldots, \textcolor{orange}{e_a^m}$ are \textit{m} element attributes, $\textcolor{yellow}{e_c^1}, \textcolor{yellow}{e_c^2}, \ldots, \textcolor{yellow}{e_c^n}$ are \textit{n} element contents (e.g., candidate texts, product images),  $\textcolor{red}{O}$ is the required output format.
Note that any layout task’s background and elements have attributes, but they do not necessarily have content.
Therefore, $\textcolor{cyan}{b_c}$  and each  $\textcolor{yellow}{e_c^i}$ may  be empty.
For clarity, we provide an instruction example for the BCEC task below, where the underlined segments correspond to their counterparts in Equation (\ref{eq:instruct}):
\begin{tcolorbox}[colback=gray!10, colframe=gray!30, title=\textbf{Instruction Example}, boxrule=0.5pt, arc=2pt, left=4pt, right=4pt, top=4pt, bottom=4pt]
\small
\setulcolor{green}
\setul{0.5ex}{0.27ex}
\ul{Task: Create an engaging, product-focused layout using provided selling point elements.} 
\setulcolor{blue}
\ul{Canvas size: 513 $\times$ 750 pixels.} 
\setulcolor{cyan}
\ul{Background image: please see the given image.}
\setulcolor{orange}
\ul{Element types: text, title, and logo.} 
\setulcolor{yellow}
\ul{Selling point candidates: ["50\% Off Today", "Shop Now", "Soft \& Breathable"].} 
\setulcolor{red}
\ul{Output format: $\text{`element\_type'}: [x_{\text{min}}, y_{\text{min}}, x_{\text{max}}, y_{\text{max}}]$.}
\end{tcolorbox}

\subsubsection{Model Training}
The unified generator ($\mathrm{G}_{\Theta}$) is trained using the next-token prediction objective across all tasks. 
Let $L$ denotes the length of $l$.
The training objective is to minimize the negative log-likelihood loss for generating the target layout $l$, given the visual input $\mathbf{I}_\mathrm{v}$ and instruction $\mathbf{X}_\mathrm{instruct}$:
\begin{equation}
\mathcal{L}_\mathrm{Gen} = - \sum_{i=1}^{L} \log p_{\boldsymbol{\Theta}}(l^{[i]}|\mathbf{I}_\mathrm{v},\mathbf{X}_\mathrm{instruct},l^{[1:i-1]}),
\label{eq:loss}
\end{equation}
\noindent where $\boldsymbol{\Theta}$ are the parameters of the generator $\mathrm{G}_{\Theta}$, and $l^{[1:i-1]}$ represents the sequence of previously generated tokens.

\subsection{Human-Mimicking Evaluation}\label{subsec:uni_evaluation}
\subsubsection{Dual-Branch Learning Strategy}\label{subsubsec:dual}
We propose a dual-branch processor $\mathrm{D}_{\Omega}$ to better simulate human evaluation of layouts, as shown in Figure~\ref{fig:outline_overall}(b). It takes the layout prediction $l$ and $\mathbf{I}_\mathrm{v}$ as inputs and generates a more human-like representation $(l_{\mathrm{d}}, \mathbf{I}_{\mathrm{d}})$, which can be formally represented as $(l_{\mathrm{d}}, \mathbf{I}_{\mathrm{d}}) \leftarrow \mathrm{D}_{\Omega}(l, \mathbf{I}_\mathrm{v})$. Specifically, the dual-branch learning strategy involves:
\begin{itemize}[nosep, leftmargin=*]
    \item \textbf{Visualization Branch:} This branch begins by visualizing different element types using distinct color blocks on the background image (or on a blank canvas for the BFEF task). 
    For the BFEC and BCEC tasks, we further display the corresponding element content to obtain the visually enhanced image $\mathbf{I}_{\mathrm{d}}$.
    This branch enhances the model's ability to capture both spatial and semantic relationships in ways tailored to each specific task, thereby better mimicking human visual perception.

    \item \textbf{Geometry Branch:} Simultaneously, the geometry branch extracts detailed spatial coordinates from the predicted layout $l$, encoding information such as element positions and sizes. It also incorporates color information from the visualization branch, creating an enriched prompt input $l_{d}$. This enriched representation not only provides a quantitative structural evaluation but also complements the visual representation.
\end{itemize}

\subsubsection{Quantitative Evaluation with CoT Reasoning}\label{cot}

Building on the dual-branch architecture, we propose a MLLM based evaluation framework that combines quantitative classification with CoT reasoning to assess layout quality. This approach not only ensures precise decision-making but also enhances interpretability by mimicking human designers' evaluation processes.

Specifically, we introduce a quantitative classification that directly predicts layout quality labels. 
In line with common practices for sequence classification using MLLMs~\citep{liu2023visual}, we utilize the hidden state of the last token $h \in \mathds{R}^d$ as the discriminative representation, capturing the cumulative contextual information of the entire input sequence. Next, a classification head $FC_{cls}$ maps the final token's hidden state $h$ to a two-dimensional probability distribution $p \in \mathds{R}^2$, which serves as the confidence estimation for layout classification:
\begin{equation}
p = \mathrm{softmax}(FC_{cls}(h)).
\label{eq:cls_head}
\end{equation}

The CoT reasoning mechanism aids layout quality assessment by bridging automated evaluation and human reasoning. As shown in Figure~\ref{fig:outline_overall}(c), our CoT mechanism analyzes layouts in four stages:
\begin{enumerate}[nosep, leftmargin=*]
    \item \textbf{Layout Glimpse:} The evaluation starts with a quick yet comprehensive scan of $\mathbf{I}_d$, capturing the layout's first impression through a concise textual description that outlines the overall composition and contextual elements.
    \item \textbf{Spatial Deconstruction:} This stage systematically breaks down the layout into its fundamental components, analyzing geometric properties and spatial relationships. It examines alignment patterns, identifies potential overlaps, and evaluates spacing consistency to uncover the underlying structural framework.
    \item \textbf{Aesthetic Appraisal:} A detailed evaluation of the layout's visual qualities is performed, focusing on artistic merit and design principles. This includes assessment of proportional balance, spatial harmony, and visual rhythm, while considering how these elements contribute to the overall aesthetic impact.
    \item \textbf{Holistic Evaluation:} The final stage synthesizes insights from all previous analyses to deliver a comprehensive assessment of the layout’s effectiveness, concluding with a clear judgment of “pass” or “fail.”
\end{enumerate}


    
        

        
        
        


\subsubsection{Model Training}
We use advanced (M)LLMs to create ground truth for CoT data. Our method starts with GPT-4o~\citep{achiam2023gpt} generating detailed captions in Stage 1. Using these captions, layout details, and human-evaluated conclusions, DeepSeek-R1~\citep{guo2025deepseek} produces precise ground truth in Stages 2 and 3. In Stage 4, we add human-annotated layout quality labels as the final conclusion. To train our evaluator for expert-like reasoning, we use curated CoT data with autoregressive next-token prediction loss ($\mathcal{L}_{\text{CoT}}$).
In addition, the $FC_{cls}$  in Equation (\ref{eq:cls_head}) is simultaneously optimized using the binary cross-entropy loss $\mathcal{L}_{\mathrm{CE}}$~\citep{ho2019real}.

Finally, the overall training objective is formulated as a combined loss function:
\begin{equation}
\mathcal{L}_{\mathrm{Eval}} = \mathcal{L}_{\text{CE}} + \mathcal{L}_{\text{CoT}}.
\end{equation}
Combining $\mathcal{L}_{\text{CE}}$ and $\mathcal{L}_{\text{CoT}}$, the evaluation loss $\mathcal{L}_{\mathrm{Eval}}$ optimizes the human-mimicking evaluator $\mathrm{E}_{\Phi}$  by ensuring a comprehensive assessment that takes into account both the accuracy of the final prediction and the quality of the reasoning process.

\subsection{Alignment with Human Feedback}\label{subsec:alignment}
We formulate layout generation as a preference-based optimization problem, aiming to guide our unified generator towards producing visually appealing layouts ($l^{+}$) while avoiding unattractive ones ($l^{-}$). Our proposed Dynamic-Margin Preference Optimization (DMPO) integrates layout generation and evaluation into a unified feedback loop, consisting of two key steps: (1) generating and evaluating pairs of candidate layouts, and (2) fine-tuning the generator based on real-time evaluator feedback. 

Given an input prompt $\mathbf{X_{instruct}}$ and a conditional background or element image $\mathbf{I_{v}}$, the generator produces two candidate layouts, $l^1$ and $l^2$.  While CoT reasoning is required during training, at inference time the evaluator utilizes only its classification head to determine the superior ($l^{+}$) and inferior ($l^{-}$) layouts, efficiently providing preference signals for generator fine-tuning.

Current preference learning methods, which often use direct likelihood maximization or fixed margin-based objectives, fail to account for the varying strengths of human judgments by treating all preference pairs equally.
To address this limitation, we propose \textit{DMPO}, which introduces a dynamic confidence margin to better quantify the relative quality differences between layout pairs. Formally, let $E_{\Phi}(l^k_{d}, \mathbf{I}^k_{\mathrm{d}})$ denotes the evaluator's score for $D_{\Omega}(l^k, I_v)$. We define the confidence margin $\delta$ between the preferred layout $l^+$ and the less preferred layout $l^-$ as:
\begin{equation}
\label{eq:margin}
\delta = E_{\Phi}(l^+,I^+) - E_{\Phi}(l^-,I^-), \quad \delta \in (0, 1],
\end{equation}

\noindent
where $I^+, I^-$ represents the images corresponding to  $l^+, l^-$, respectively.
To further amplify the perception of the margin, we apply a non-linear transformation, given by \( f(\delta) = e^{\delta} - e^{-\delta} \), to the confidence margin.
Finally, the DMPO loss function is formulated as follows:
{\normalsize
\begin{align}
\label{eq:DMPO}
\mathcal{L}_{\text{DMPO}} = & -\log\sigma\bigg(\beta \log \frac{G_{\Theta}(l^+|I_v, X_{instruct})}{G_{ref}(l^+|I_v, X_{instruct})}  \nonumber \\ 
& - \beta \log \frac{G_{\Theta}(l^-|I_v, X_{instruct})}{G_{ref}(l^-|I_v, X_{instruct})} - f(\delta)\bigg),
\end{align}
}
\noindent
where $\sigma(\cdot)$ is the sigmoid function, $\beta$ controls optimization sensitivity, and $G_{\Theta}(l|I_v,  X_{instruct})$ represents the layout generation probability given image $I_v$ and instruction $X_{instruct}$. 
By integrating generation and evaluation in a feedback loop, DMPO bridges the gap between layout generation and human aesthetic preferences, producing more visually appealing layouts.

\begin{table}[t]
\centering
\caption{Comparison of accuracy (\%) across different methods on LayoutHF-100k. Bold and underlined denote the best and second-best results, respectively (the same below). }
\resizebox{\columnwidth}{!}{%
\begin{tabular}{l|cccc|c}
\toprule
Model & BFEF & BCEF & BFEC & BCEC & Overall \\ 
\midrule
Claude3.5~\citep{claude2024} & 50.7 & \underline{66.8} & 51.9 & 61.9 & 57.8 \\ 
DeepSeek-R1~\citep{guo2025deepseek} & 54.7 & 60.9 & 52.0 & 49.6 & 54.3 \\
GLM-4v~\citep{glm2024chatglm} & 50.5 & 48.5 & 47.4 & 47.6 & 48.5 \\ 
GPT-4o~\citep{achiam2023gpt} & \underline{58.3} & 64.9 & \underline{60.8} & \underline{62.3} & \underline{61.6} \\ 
\rowcolor[HTML]{EFEFEF} 
Ours & \textbf{86.2} & \textbf{87.2} & \textbf{88.2} & \textbf{80.4} & \textbf{85.5} \\ 
\bottomrule
\end{tabular}%
\label{tab:reward-sota}
}
\end{table}

\section{Experiment} 
\subsection{Implementation Details}. 
In this paper, we use LLaVA~\citep{liu2023visual} as the base model for the generator and the evaluator, respectively. For the unified generator pre-training phase, we conduct full model fine-tuning over 10 epochs using a cosine learning rate scheduler with an initial learning rate of 2e-6. This phase takes approximately 5 days to complete. Subsequently, we initialize the evaluator with these pre-trained weights  and apply the same learning strategy to train on \textit{Layout-HF100k} dataset, which simulates human preferences. 
In the final DMPO stage, we leverage the frozen evaluator to guide our proposed \textit{Uni-Layout} generation framework. During this phase, we implement LoRA~\citep{hu2022lora} fine-tuning with a learning rate of 2e-5, running for 3 epochs, which takes about 50 hours to complete. All experiments are conducted on a single node with 8 NVIDIA H100 GPUs.

\subsection{Layout Evaluation Performance}
\subsubsection{Evaluation Metric}

To assess the human-mimicking evaluator's effectiveness, we define the \textit{accuracy} metric as follows:
\begin{equation}
\text{accuracy} = \frac{1}{N} \sum_{i=1}^N \mathds{1}(\hat{y_i} = y_i),
\end{equation}
\noindent where \(N\) represents the total number of layout predictions, \(\hat{y_i}\) is the predicted label, \(y_i\) is the true label, and \(\mathds{1}\) is the indicator function.

\subsubsection{Comparison with SOTA}
To validate our  evaluator, we compare it with leading closed-source (M)LLMs, including GPT-4o~\citep{achiam2023gpt}, Claude3.5 Sonnet~\citep{claude2024} (Claude3.5), GLM-4v~\citep{glm2024chatglm}, and DeepSeek-R1~\citep{guo2025deepseek}. These models follow the "LLM-as-Judge"~\citep{zheng2023judging} paradigm. All models receive identical instructions and visual inputs, except DeepSeek-R1, which only processes text. Outputs are converted to predicted labels for evaluation; our model uses a classification head instead of CoT reasoning for efficient label prediction. As shown in Table~\ref{tab:reward-sota}, our model consistently excels, achieving 85.5\% accuracy, outperforming existing MLLMs by 25-35\%. Some MLLMs perform near-randomly ($\approx$50\%), highlighting their limitations in layout evaluation. 
Our \textit{Layout-HF100k} dataset fills this gap and supports future model development.
These results underscore the significant gap in layout evaluation. Our \textit{Layout-HF100k} dataset not only bridges this critical void but also lays a robust foundation for the development and training of future layout evaluation models.

\begin{table}[t]
\centering
\caption{Ablation study for the human-mimicking evaluator.}
\resizebox{\columnwidth}{!}{
\begin{tabular}{ccccc|c}
\toprule
\textbf{Separate} & \textbf{Joint} & \textbf{Visualization} & \textbf{Classification} & \textbf{CoT} & \textbf{accuracy (\%)} \\
\midrule
\ding{55} & \ding{55} & \ding{55} & \ding{55} & \ding{55} & 50.6 \\
\ding{51} & \ding{55} & \ding{55} & \ding{55} & \ding{55} & 77.8\textcolor{blue}{ (+27.2\%)} \\
\ding{51} & \ding{51} & \ding{55} & \ding{55} & \ding{55} & 79.9\textcolor{blue}{ (+2.1\%)} \\
\ding{51} & \ding{51} & \ding{51} & \ding{55} & \ding{55} & 82.5\textcolor{blue}{ (+2.6\%)} \\
\ding{51} & \ding{51} & \ding{51} & \ding{51} & \ding{55} & 83.1\textcolor{blue}{ (+0.6\%)} \\
\ding{51} & \ding{51} & \ding{51} & \ding{51} & \ding{51} & 85.5\textcolor{blue}{ (+2.4\%)} \\
\bottomrule
\end{tabular}
\label{tab:ablation_reward}
}
\end{table}

\begin{table*}[h]
\centering
\caption{Comparison of public metrics across different methods and tasks. Models marked with * are our own implementations. From top to bottom, the models are organized as task-specific, general-purpose and ours.}
\resizebox{0.95\textwidth}{!}{%
\begin{tabular}{l|ccc|ccc|ccc|ccc}
\toprule
Tasks & \multicolumn{3}{c|}{BFEF} & \multicolumn{3}{c|}{BFEC} & \multicolumn{3}{c|}{BCEF} & \multicolumn{3}{c}{BCEC} \\
Metrics & Ove↓ & Ali↓ & Max.↑ & Ove↓ & Ali↓ & Max.↑ & $R_{com}$↓ & $R_{sub}$↓ & $R_{occ}$↑ & $R_{com}$↓ & $R_{sub}$↓ & $R_{occ}$↑ \\
\midrule
LayoutDM~\citep{inoue2023layoutdm} & 0.139 & 0.267 & \underline{0.348} & - & - & - & - & - & - & - & - & - \\
LayoutFlow~\citep{guerreiro2024layoutflow} & 0.011 & 0.037 & \textbf{0.350} & - & - & - & - & - & - & - & - & - \\
P\&R*~\citep{li2023planning} & - & - & - & 0.054 & 0.015 & \underline{0.365} & - & - & - & - & - & - \\
PAID*~\citep{chen2025paid} & - & - & - & 0.209 & 0.007 & 0.178 & - & - & - & - & - & - \\
CGL-GAN~\citep{zhou2022composition} & - & - & - & - & - & - & 34.010 & 0.816 & 0.997 & - & - & - \\
PosterLlama~\citep{seol2024posterllama} & - & - & - & - & - & - & 42.722 & 0.804 & 0.990 & 28.480 & 0.774 & 0.961 \\
RADM~\citep{li2023relation} & - & - & - & - & - & - & - & - & - & \underline{10.260} & \textbf{0.742} & \underline{0.997} \\
\midrule
GPT-4o~\citep{achiam2023gpt} & \textbf{0.001} & \underline{0.001} & 0.074 & \underline{0.002} & 0.005 & 0.239 & \underline{32.814} & 0.794 & \textbf{1} & 27.168 & 0.792 & \textbf{1} \\
DeepSeek-R1~\citep{guo2025deepseek} & 0.085 & 0.071 & 0.022 & 0.078 & \underline{0.004} & 0.320 & 39.778 & 0.813 & \textbf{1} & 27.012 & 0.885  & \textbf{1} \\
Claude3.5~\citep{claude2024} & 0.008 & \underline{0.001} & 0.091 & 0.005  & \textbf{0.001} & 0.242 & 40.819  & \underline{0.779} & \textbf{1} & 31.157 & 0.773 & \textbf{1} \\
LLaVA~\citep{liu2023visual} & \underline{0.002} & 0.014 & 0.073 & 0.006 & 0.018 & 0.300 & 42.472 & 0.782 & \underline{0.998} & 22.813 & 0.785 & 0.948 \\
\midrule
\rowcolor[HTML]{EFEFEF}
Ours & \textbf{0.001} & \textbf{0.00004} & 0.160 & \textbf{0.00045} & 0.009 & \textbf{0.439} & \textbf{31.848} & \textbf{0.774} & \textbf{1} & \textbf{8.536} & \underline{0.764} & \textbf{1} \\
\bottomrule
\end{tabular}%
\label{tab:generation}
}
\end{table*}

\subsubsection{Ablation Studies}
In Table~\ref{tab:ablation_reward}, we present a series of ablation experiments conducted on the Layout-HF100k dataset to evaluate the contribution of each component to the evaluator.

\begin{itemize}[nosep, leftmargin=*]

    \item \textbf{Separate Training:} The baseline model employs a basic LLaVA and lacks layout evaluation capabilities. By adopting a separate training approach, 
    the accuracy significantly improves by 27.2\%, reaching 77.8\%. This underscores the importance of a high-quality dataset and effective pre-training for layout evaluation.

    \item \textbf{Joint Training:} The joint training strategy involves training a single LLaVA model simultaneously across all four datasets, which further increases the accuracy by 2.1\% to 79.9\%. 

    \item \textbf{Visualization:} Incorporating visualization through a dual-branch learning strategy adds another 2.6\% improvement, reaching 82.5\%. This highlights the crucial role of visual information in multimodal layout evaluation.

    \item \textbf{Classification Head:} The introduction of a classification head, which directly generates numerical labels instead of textual predictions, achieves a 0.6\% performance improvement to 83.1\%. 

    \item \textbf{CoT Reasoning:} Finally, augmenting the model with a CoT reasoning mechanism results in a 2.4\% improvement, achieving a final accuracy of 85.5\%. This demonstrates the significant impact of enhanced reasoning capabilities in layout evaluation tasks.
\end{itemize}

\begin{figure}[t]

    \centering
    \includegraphics[width=0.95\columnwidth]{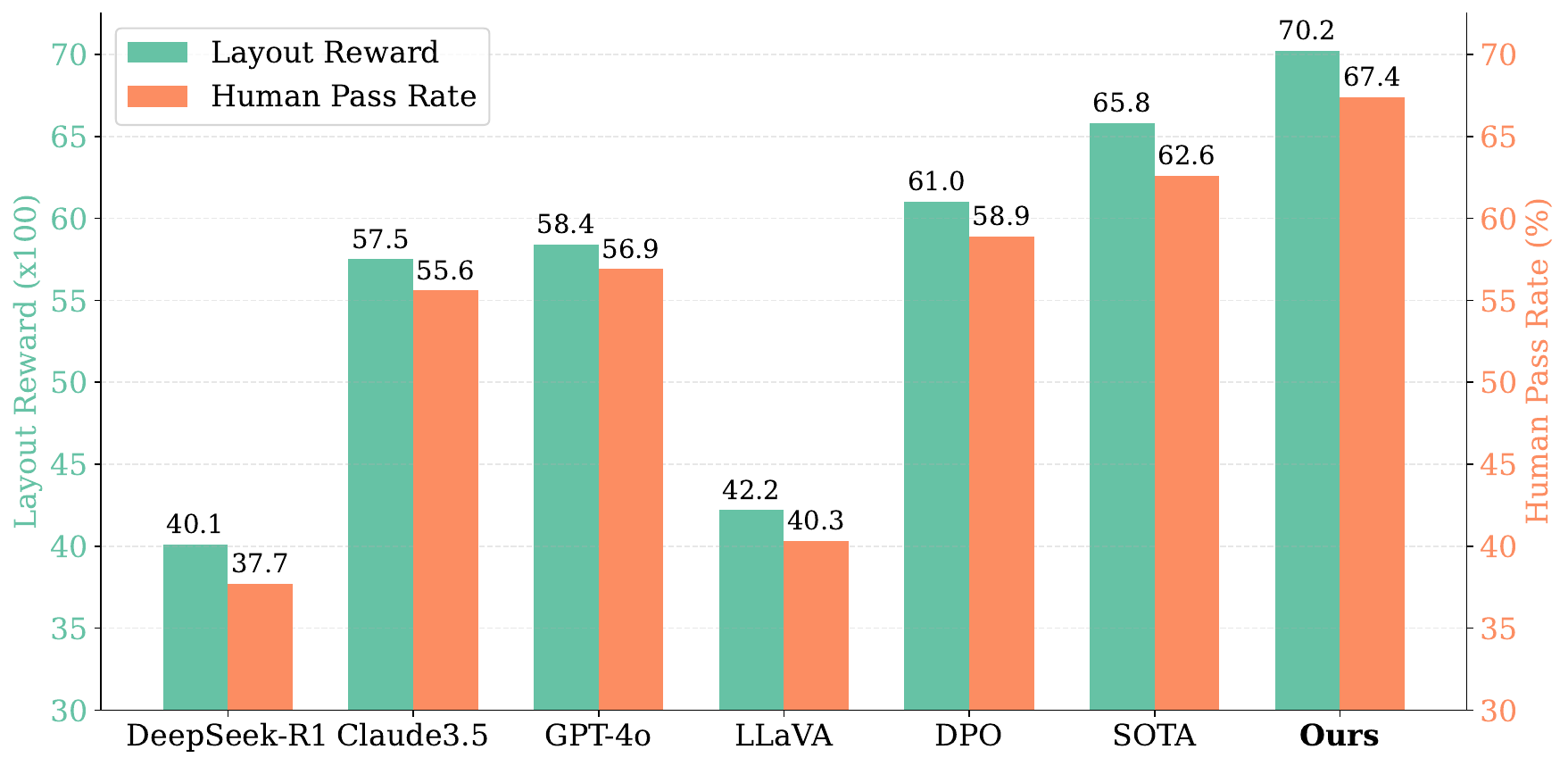}
    \caption{Layout Reward  ($\times 100$) and Human Pass Rate (\%)  across different methods.}
    \label{fig:reward_scores}

\end{figure}

\subsection{Layout Generation Performance}

\subsubsection{Evaluation Metric}\label{subsubsec:metric}
To evaluate the quality of layout generation, we employ both task-specific and our proposed human-mimicking evaluation metrics. 

For the task-specific evaluation, we adopt metrics tailored to each task's requirements. For the BFEF and BFEC tasks, which focus on geometric relationships, we utilize metrics such as Overlap (Ove)~\citep{hsu2023posterlayout}, Alignment (Ali)~\citep{li2020attribute}, and Maximum IoU (Max.)~\citep{kikuchi2021constrained}. 
For the BCEF and BCEC tasks, which require background image perception, we apply composition-relevant metrics like $R_{com}$, $R_{sub}$, and $R_{occ}$~\citep{zhou2022composition, li2023relation}, focusing on readability, product presentation, and layout occupancy.

For the human-mimicking evaluation, we utilize the human-mimicking evaluator to obtain the \textit{Layout Reward (LR)}, a  score derived from the classification head of our evaluator. Specifically, we use the positive class probability $p[1]$ in Equation (\ref{eq:cls_head}), yielding a score between 0 and 1. Formally, LR can be expressed as: 
\begin{equation}
\text{LR} = p[1].
\end{equation} 

\subsubsection{Comparison with SOTA}
We conduct comprehensive comparisons with three categories of baseline methods: (1) task-specific SOTA models designed for individual layout tasks (e.g., LayoutDM~\citep{inoue2023layoutdm}); (2) closed-source models including GPT-4o~\citep{achiam2023gpt}, Claude3.5~\citep{claude2024}, and DeepSeek-R1~\citep{guo2025deepseek} and (3) open-source MLLMs like LLaVA that have been jointly trained on four tasks. 


    
    
    
For task-specific evaluations shown in Table~\ref{tab:generation}, our method demonstrates outstanding performance across multiple metrics. Notably, in the BFEF task, we achieve the lowest \textit{Ove} (0.001) and \textit{Ali} (0.00004), matching or surpassing specialized models such as LayoutDM and LayoutFlow. In the BFEC task, our approach sets a new state-of-the-art with minimal \textit{Ove} (0.00045) and the highest Max. (0.439). For the BCEF task, we attain the best results in $R_{com}$ (31.848) and $R_{sub}$ (0.774). Similarly, in the BCEC task, our method significantly outperforms existing methods with the lowest $R_{com}$ (8.536), while keeping competitive performance in other metrics. 
Like closed-source (M)LLMs, our method achieves perfect occupancy ($R_{occ}$ = 1) with no empty predictions.
Overall, our approach consistently achieves top-tier results in key metrics and remains close to the best in others, demonstrating its robustness and effectiveness across diverse tasks.

For human-mimicking evaluation across all tasks, we incorporate the LR score to assess the performance. As shown in  Figure~\ref{fig:reward_scores}, our method achieves the highest LR score of 0.702, demonstrating consistent superiority across different layout scenarios. Compared to other models, we significantly outperform GPT-4o (0.584), Claude3.5 (0.575), and DeepSeek-R1 (0.401) by substantial margins. The performance gap is even more pronounced when compared to the open-source baseline LLaVA (0.422), showing an improvement of nearly 30\%. 
Compared to the average LR score of 0.658 achieved by LayoutFlow (SOTA for BFEF), P\&R (SOTA for BFEC), and PosterLlama (SOTA for BCEF and BCEC), our approach achieves superior results, thereby validating the effectiveness of \textit{Uni-Layout}.

\subsubsection{Human Evaluation}
The Human Pass Rate (HPR) represents the percentage of generated layouts that successfully pass human evaluation. As shown in  Figure~\ref{fig:reward_scores}, our method achieves the highest HPR at 67.4\%, outperforming GPT-4o (56.9\%), Claude3.5 (55.6\%), and DeepSeek-R1 (37.7\%). Compared to LLaVA (40.3\%), our approach shows an improvement of nearly 30\%. Notably, we surpass the previous SOTA (62.6\%) by 4.8\%, further demonstrating the effectiveness of our approach. Furthermore, the trend in HPR closely aligns with LR validating the effectiveness of the proposed LR.

\subsubsection{Ablation Studies}

\begin{figure}
    \centering
    \includegraphics[width=0.9\columnwidth]{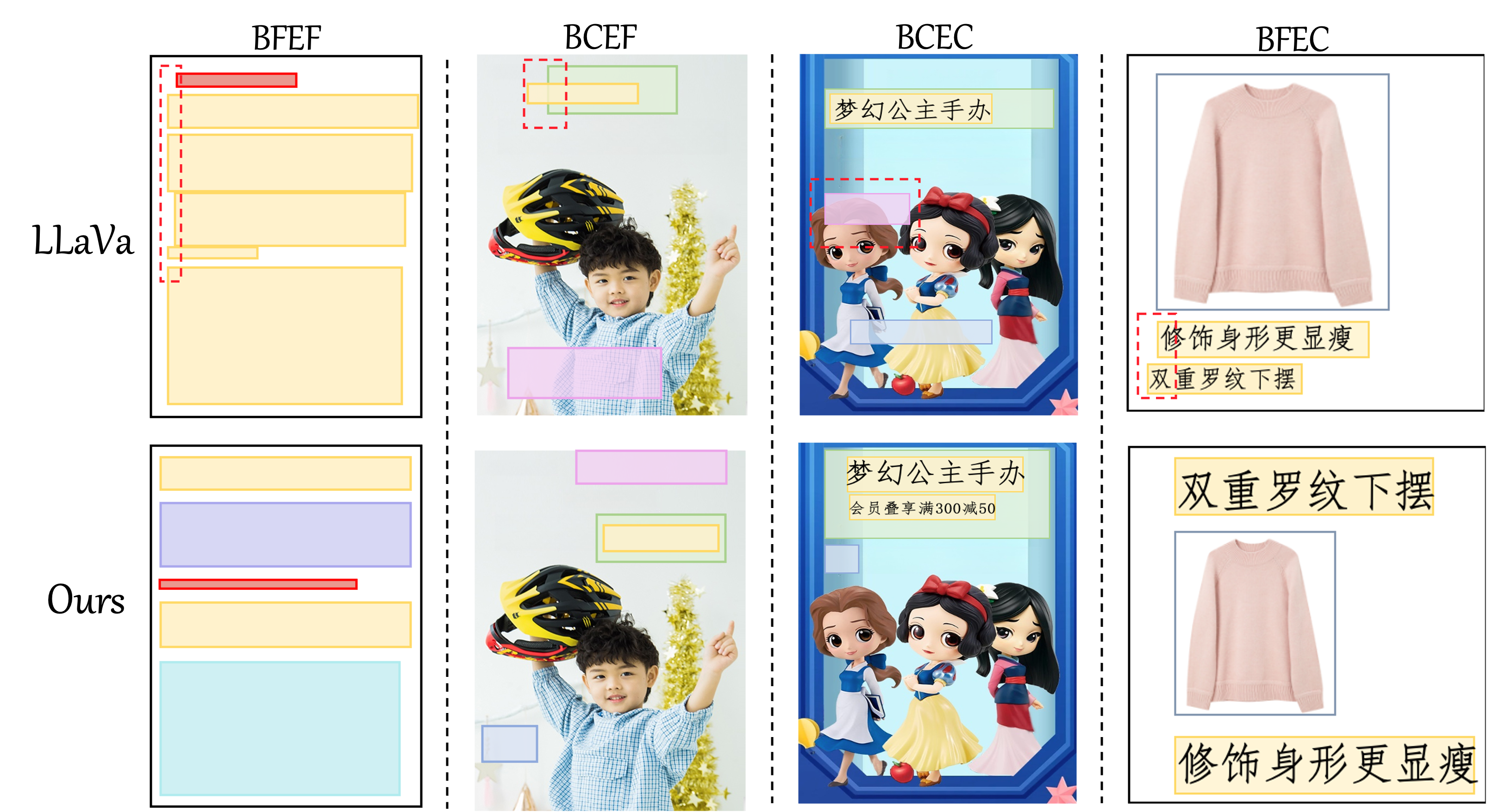}
    \caption{Comparison of effects before and after alignment.}
    \label{fig:cpm-alinment}
\end{figure}

To explore the contributions of our key components, we conduct ablation studies on two aspects.:
\begin{itemize}[nosep, leftmargin=*]
    \item \textbf{Effectiveness of Alignment}: By integrating human feedback with DMPO, \textit{Uni-Layout} achieves superior performance as evidenced by both quantitative metrics (Table~\ref{tab:generation}) and human evaluation results (Figure~\ref{fig:reward_scores}).  
    The visualization results in Figure~\ref{fig:cpm-alinment}  further intuitively demonstrate the effectiveness of alignment.
    \item \textbf{Superiority of DMPO:} We evaluate the dynamic-margin mechanism by comparing DMPO with DPO and fixed margin methods. 
    DMPO achieves a higher LR score of 0.702, outperforming DPO's 0.610.
    In comparison to fixed margins of 0.5 (LR=0.625), 1 (LR=0.667), 1.5 (LR=0.674) and 2 (LR=0.658), DMPO consistently outperforms these settings.
    These results confirm that DMPO helps generate  human-preferred layouts.
\end{itemize}

\section{Conclusion}
In this paper, we introduce \textit{Uni-Layout}, a unified framework that seamlessly integrates layout generation, evaluation and alignment, addressing the limitations of existing task-specific approaches. Our framework features a unified generator capable of handling diverse layout tasks via natural language prompts and a novel human-mimicking evaluator trained on \textit{Layout-HF100k}, a large-scale human feedback dataset that we created. To further enhance alignment with human preferences, we propose Dynamic-Margin Preference Optimization (DMPO), which dynamically adjusts margins based on preference strength. Extensive experiments across universal layout generation tasks demonstrate that \textit{Uni-Layout} significantly outperforms both task-specific and general-purpose methods, achieving state-of-the-art results in both task-specific and human-mimicking evaluation metrics, including HPR and LR score. Our work highlights the importance of human feedback in layout generation and paves the way for future research in this area. 

\section{Acknowledgments}
This work was supported by the National Natural Science Foundation of China (62276256, U2441251), the Young Elite Scientists Sponsorship Program by CAST (2023QNRC001), and the Young Scientists Fund of the State Key Laboratory of Multimodal Artificial Intelligence Systems (ES2P100117).

\bibliographystyle{ACM-Reference-Format}
\balance
\bibliography{main}

\clearpage
\newpage
\appendix

\section{Appendix}
In the appendix, we provide supplementary materials to support research reproducibility. We include detailed annotation protocols  used in constructing the Layout-HF100k dataset. Additionally, we present visual demonstrations of both exemplary and challenging cases to highlight our model’s capabilities and areas for improvement. The appendix also discusses current limitations, future research directions, and societal implications. This documentation underscores our commitment to transparency in advancing automatic layout generation.

\subsection{Annotation Guidance}
\label{sec:anno_guidance}
Annotators evaluate layouts individually, with different evaluation criteria based on task types. For background-constrained tasks, annotators reference the background image, while element-constrained tasks include additional textual content. Annotators determine whether a layout is qualified based on our detailed guidelines that specify disqualifying characteristics. 

\subsubsection{Unqualified Criteria for the Background-Free and Element-Constrained Task}\label{a11}
\begin{itemize}
\item \textbf{Disorder.} Evaluate whether the distribution of elements within the image (such as object and text) follows an orderly pattern. If the elements are scattered randomly without clear grouping, alignment, or spacing, such as multiple objects being haphazardly piled together with chaotic distances and positional relationships.
\item \textbf{Misaligned.} Alignment can be categorized into group alignment and visual perceptual alignment, with the absence of either indicating misalignment. Specifically, group alignment is further divided into intra-group and inter-group alignment. When multiple groups of elements are present in an image, it is essential to first assess whether the elements within each group adhere to a consistent alignment method, such as left, right, or center alignment, ensuring that the group appears orderly and cohesive. Subsequently, it is crucial to evaluate whether different groups follow a consistent alignment pattern, such as aligning multiple groups along a horizontal or vertical axis, to form a larger, orderly layout structure and prevent uneven spacing between groups. Visual perceptual alignment refers to the overall harmony of the layout, where the image as a whole should convey a sense of order and balanced comfort.
\item \textbf{Empty.} Assess the uniformity of element density across the image. If certain regions are overly dense with elements while others are excessively sparse, such as one section being crowded with various elements and another nearly blank, the layout may seem imbalanced. Additionally, layouts with significant blank areas, particularly when one side of the image lacks any layout boxes.
\item \textbf{Overlap.} This refers to the situation where any two layout elements overlap in position. In this task, intersections can occur between text boxes, as well as between text boxes and product images.
\end{itemize}

\subsubsection{Unqualified Criteria for the Background-Constrained and Element-Constrained Task}\label{a12}
\begin{itemize}
\item \textbf{Overlap.} This includes two types: inter-box overlap and box-background overlap. Inter-box overlap occurs between layout elements, where only underlay boxes can fully enclose other elements. Box-background overlap involves layout elements occluding background content, such as text boxes covering facial areas in the background image.
\item \textbf{Invalid Underlay.} An underlay box is deemed invalid if it does not contain any nested layout elements.
\item \textbf{Misaligned.}  The description of the misalignment metric is consistent with that provided in Section ~\ref{a11}.
\item \textbf{Extreme-sized.} Extreme-sized boxes refer to layout boxes that are either excessively large or excessively small. An excessively large box is defined as one that covers more than one-third of the image area, while an excessively small box is one where the text inside becomes illegible. Specifically, a small box is typically defined as having an area of less than 1000 pixels and a height of less than 30 pixels.
\end{itemize}

\subsubsection{Unqualified Criteria for the Background-Constrained and Element-Free Task}
\begin{itemize}
\item \textbf{Overlap.} This includes two types: inter-box overlap and box-background overlap. Inter-box overlap occurs between layout elements, where only underlay boxes can fully enclose other elements. Box-background overlap involves layout elements occluding background content, such as text boxes covering facial areas.
\item \textbf{Invalid Underlay.} The criteria for determining invalid layouts are the same as those mentioned in Section ~\ref{a12}.
\item \textbf{Misaligned.} The criteria for misalignment are the same as those described in Section ~\ref{a11}.
\end{itemize}

\subsubsection{Unqualified Criteria for the Background-Free and Element-Free Task}
\begin{itemize}
\item \textbf{Overlap.} This refers to the area overlap between any two layout boxes, regardless of whether they are of the same type or different types. Any occurrence of such overlap is considered unqualified.
\item \textbf{Empty.} The criteria for determining whether a layout is empty are the same as those specified in Section ~\ref{a11}.
\item \textbf{Disorder.} The criteria for determining whether a layout is disordered are the same as those outlined in Section ~\ref{a12}.
\item \textbf{Misaligned.} Beyond the misalignment criteria mentioned in Section ~\ref{a11}, this task requires that text boxes adjacent to table, list, and figure maintain consistent alignment with them, while all other text boxes must be left-aligned.
\end{itemize}

\subsection{More Visualization Results}
\subsubsection{Dataset Visualization} 

In this section, we showcase representative samples from both Layout-HF100k and EP-Layout datasets to demonstrate their diversity and characteristics.

\begin{figure*}[t]
    \centering
    \includegraphics[width=0.9\textwidth]{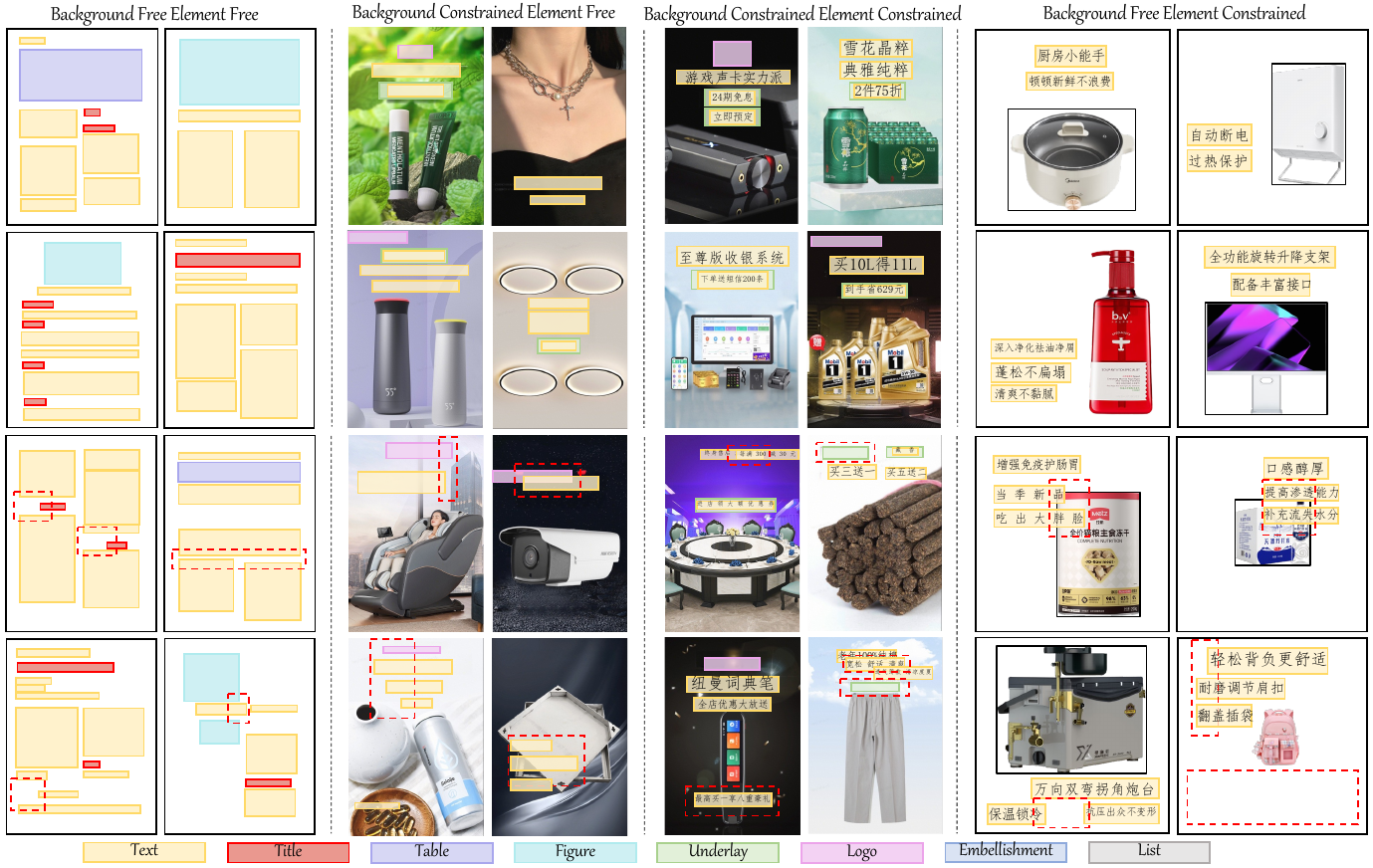}
    \caption{More examples of the proposed Layout-HF100k. The first two lines show qualified examples, while the rest shows unqualified ones. The unqualified parts are noted by the red dashed line.}    
    \label{fig:dataset_ap}
    \vspace{1em} 
    \includegraphics[width=\textwidth, height=0.5\textheight, keepaspectratio]{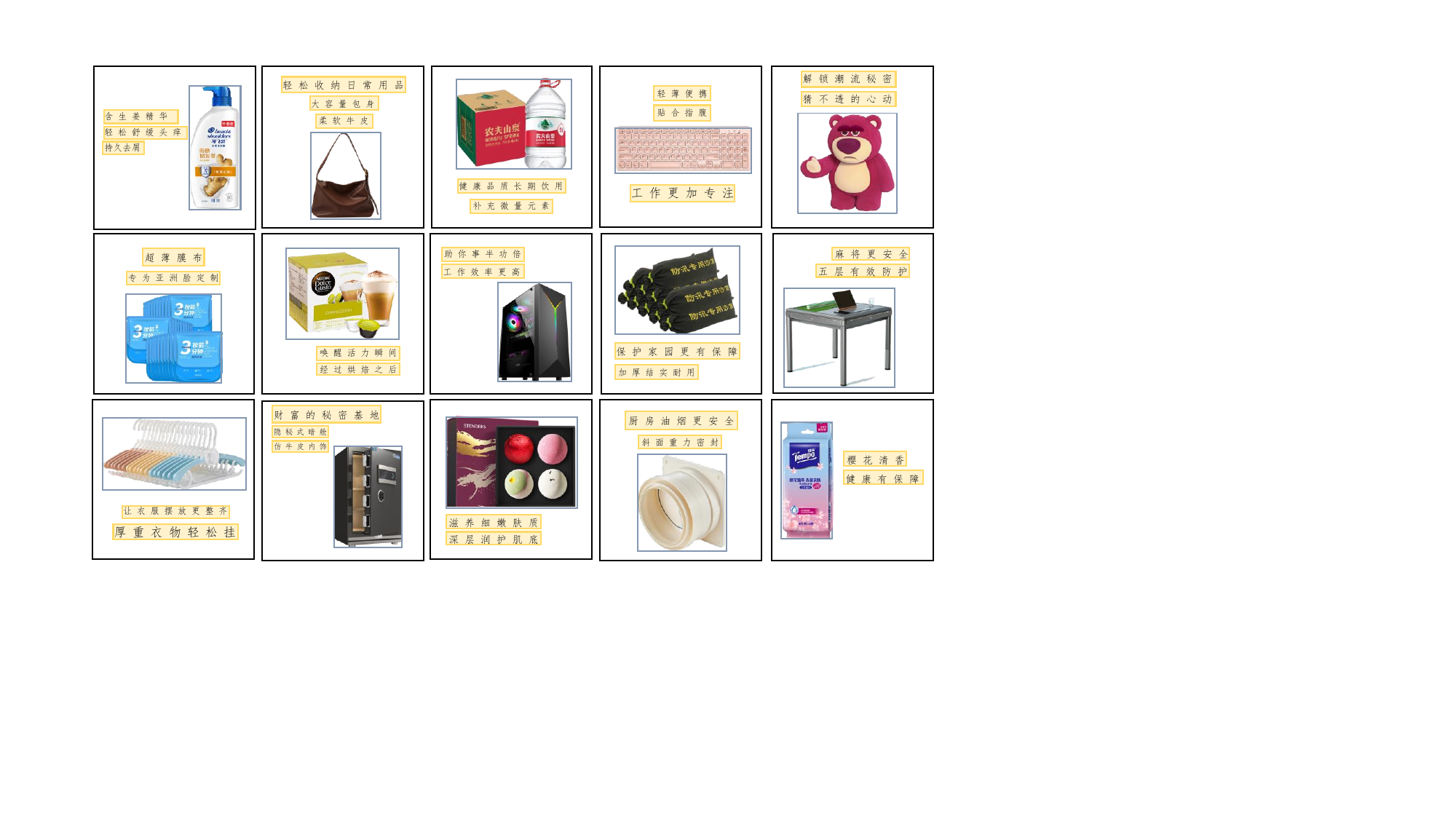}
    \caption{More examples of the proposed EP-Layout for the Background-Free and Element-Constrained Task, focusing on layout for text content and product images without background image restrictions.}
    \label{fig:dataset_ep}
\end{figure*}

Figure~\ref{fig:dataset_ap} presents Layout-HF100k examples, where the top two rows show qualified layouts with proper structure, while the bottom rows highlight unqualified cases with deficiencies marked by red dashed lines. These examples illustrate our qualification criteria.

Figure~\ref{fig:dataset_ep} displays EP-Layout samples, demonstrating the dataset’s unique features and layout variations. These examples reflect the dataset’s scope and design considerations.


\subsubsection{Comparison with State-of-the-Art Methods} 
In this section, we present visual comparisons between our Uni-Layout model and state-of-the-art (SOTA) methods across four distinct layout generation tasks. These comparisons highlight the superior performance of our model in generating diverse and high-quality layouts.

\begin{figure*}[t]
    \centering
    \includegraphics[width=0.9\textwidth]{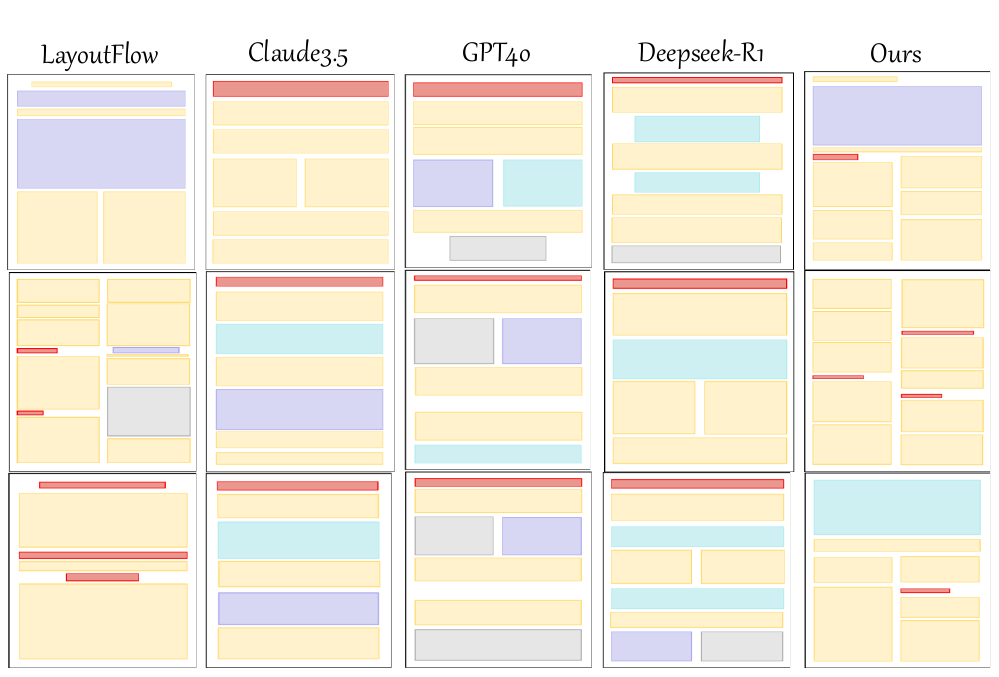}
    \caption{Comparison of different layout generation models for the Background-Free and Element-Free task.}    
    \label{fig:case_Background-Free and Element-Free}
    \vspace{1em} 
    \includegraphics[width=\textwidth, height=0.5\textheight, keepaspectratio]{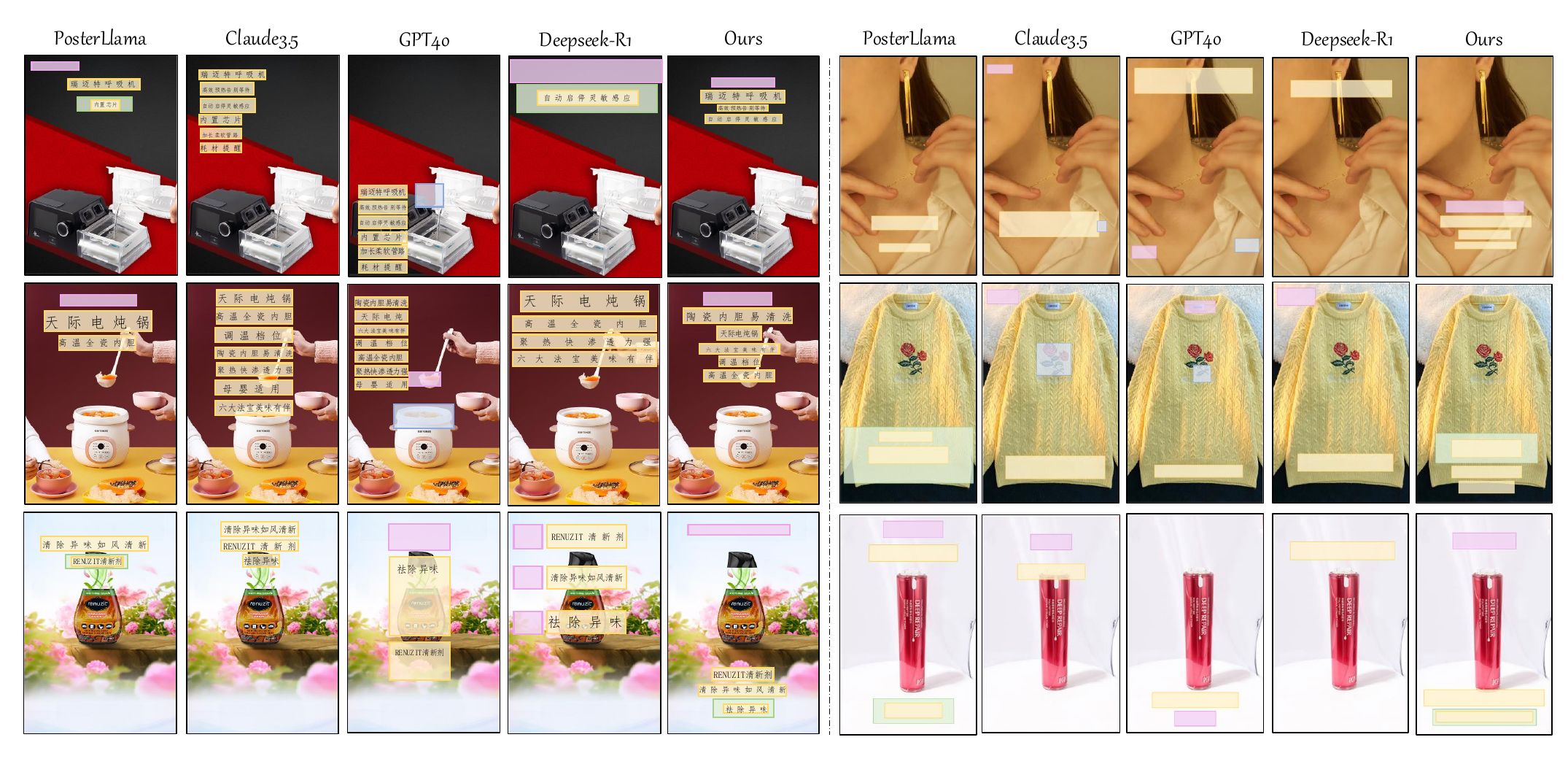}
    \caption{Comparison of different layout generation models for the Background-Constrained and Element-Constrained task (left), the Background-Constrained and Element-Free task (right).}
    \label{fig:case_Background-Constrained and Element-Constrained and case_Background-Constrained and Element-Free}
\end{figure*}

Figure~\ref{fig:case_Background-Free and Element-Free} illustrates the comparison for the Background-Free and Element-Free task. Here, our model demonstrates its capability to generate layouts without predefined backgrounds or elements, outperforming other models in terms of both creativity and adherence to layout standards.

In Figure~\ref{fig:case_Background-Constrained and Element-Constrained and case_Background-Constrained and Element-Free}, we compare models for the Background-Constrained and Element-Constrained task (left) and the Background-Constrained and Element-Free task (right). These comparisons reveal the adaptability of our model in maintaining layout coherence and quality under varying constraints, showcasing its robustness across different scenarios.

Finally, Figure~\ref{fig:case_Background-Free and Element-Constrained} focuses on the Background-Free and Element-Constrained task. The results highlight our model's proficiency in generating layouts with specific element constraints, while still achieving a high degree of aesthetic and functional quality.


\subsubsection{Alignment Effects}
In this section, we illustrate the impact of our alignment mechanism on layout quality by presenting before-and-after examples. These examples clearly demonstrate the improvements in layout coherence and precision achieved through our method.

\begin{figure*}[t]
    \centering
    \includegraphics[width=0.9\textwidth]{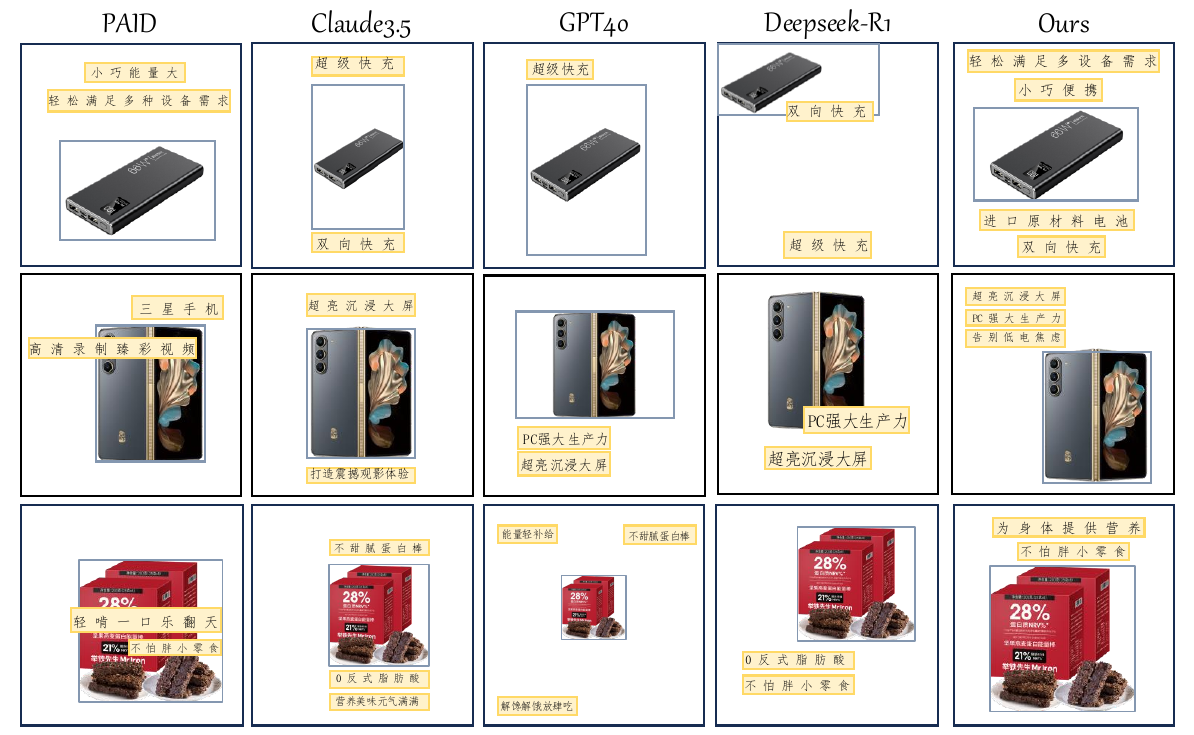}
    \caption{Comparison of different layout generation models for the Background-Free and Element-Constrained task.}           
    \label{fig:case_Background-Free and Element-Constrained}
    \vspace{1em} 
    \includegraphics[width=\textwidth, height=0.5\textheight, keepaspectratio]{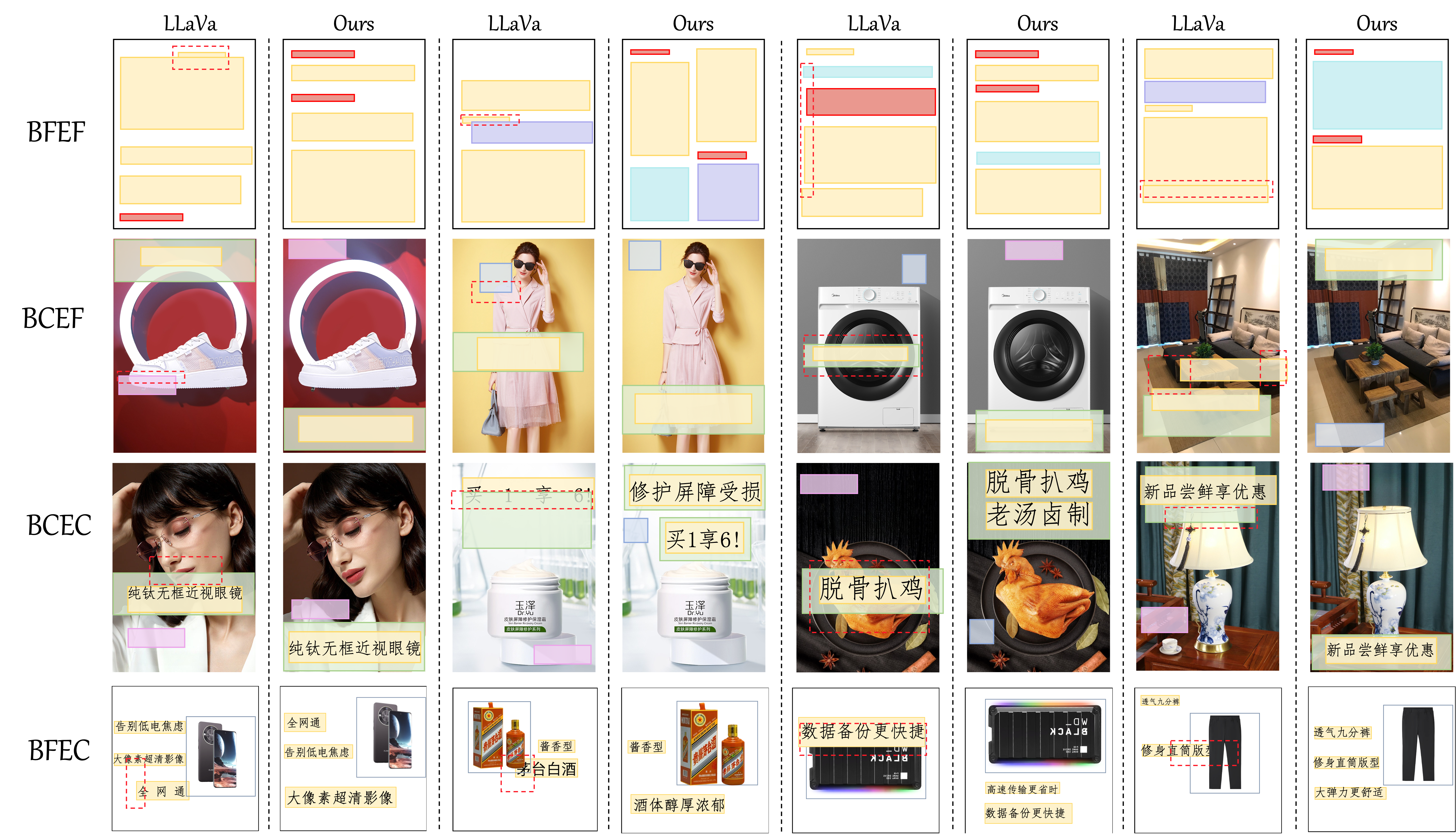}
    \caption{Comparison of effects before and after alignment. The unqualified parts are noted by the red dashed line.}
    \label{fig:cpm-all}
\end{figure*}

Figure~\ref{fig:cpm-all} showcases a comparison of layouts prior to and following the application of our alignment mechanism. The unqualified elements in the layouts, which are highlighted by red dashed lines, exhibit noticeable enhancements post-alignment. This visual evidence underscores the effectiveness of our approach in refining layout structures, ensuring that elements are well-aligned and aesthetically pleasing.

\subsubsection{Additional Generation Examples}
\begin{figure*}
    \centering
    \includegraphics[width=0.9\textwidth]{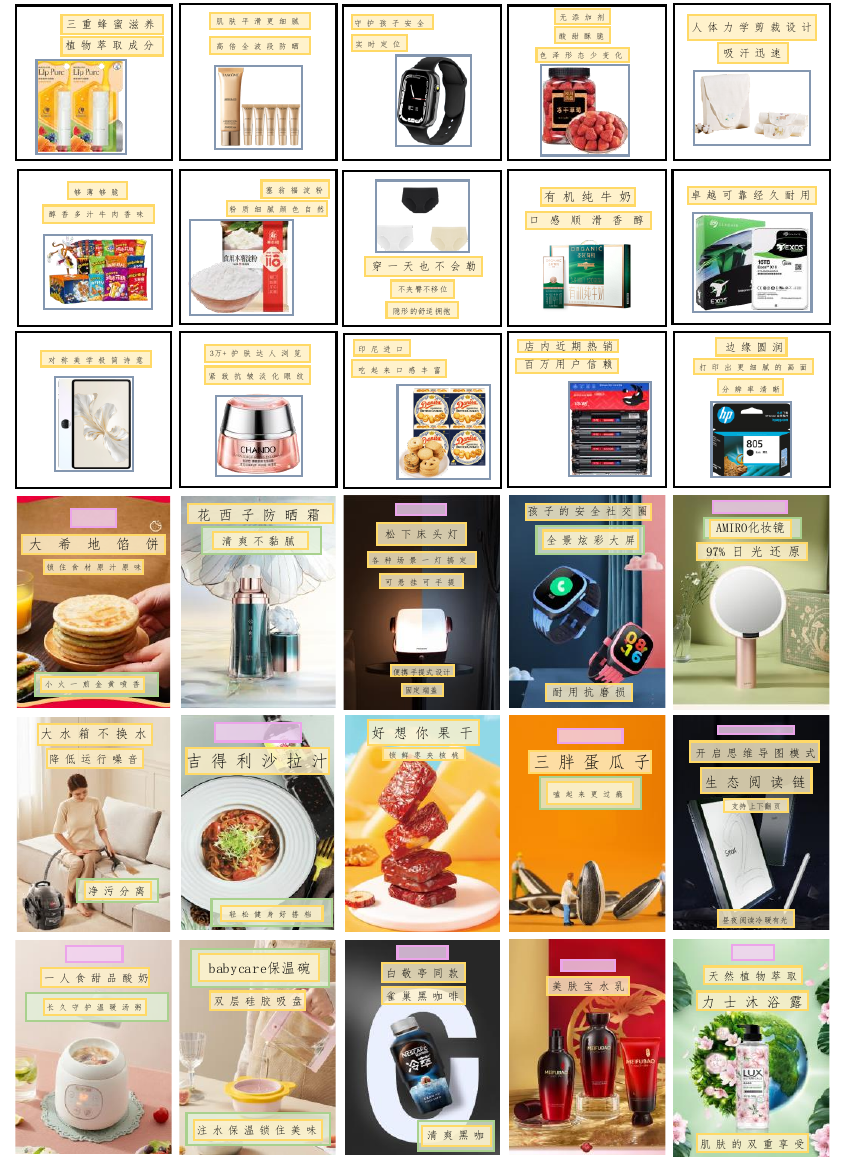}
    \caption{Layout examples generated by Uni-Layout (Part I).}           
    \label{fig:Uni-Layout-case_more1}
\end{figure*}

\begin{figure*}
    \centering
    \includegraphics[width=0.9\textwidth]{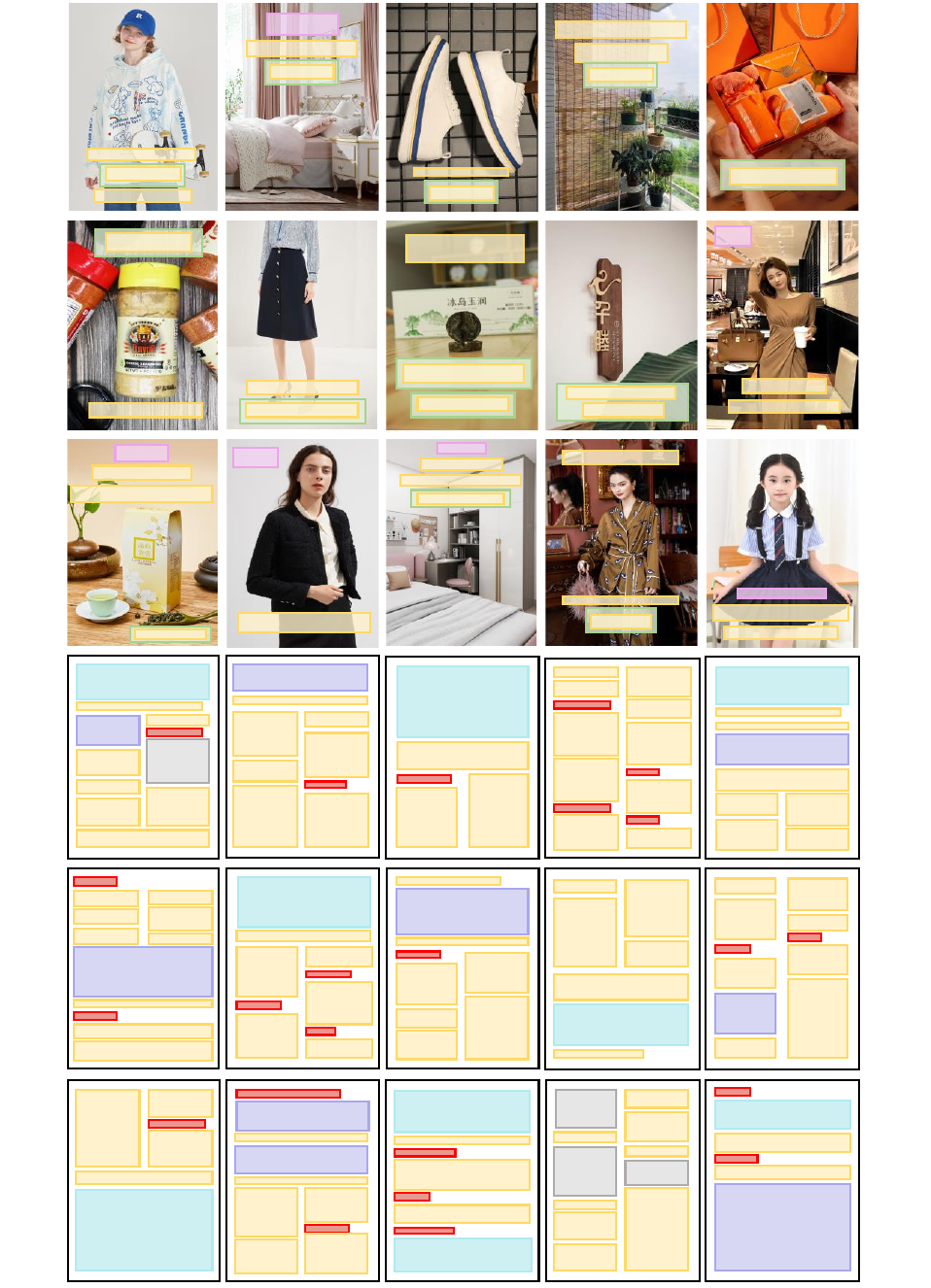}
    \caption{Layout examples generated by Uni-Layout (Part II).}           
    \label{fig:Uni-Layout-case_more2}
\end{figure*}

In this section, we present a broader array of layout generation examples produced by Uni-Layout, demonstrating its versatility and capability across various scenarios. These examples illustrate the model’s proficiency in creating diverse and high-quality layouts tailored to different requirements, as shown in Figures~\ref{fig:Uni-Layout-case_more1} and~\ref{fig:Uni-Layout-case_more2}.





\subsection{Limitations and Future work}
Our research focuses on layout for 2D graphic design. In future work, we plan to extend this work to three-dimensional space by adapting our model to 3D data. Additionally, we would build new human feedback mechanisms for 3D layout evaluation. This expansion will enable our framework to handle more complex applications, such as virtual reality (VR), augmented reality (AR), and 3D modeling.

\subsection{Social Impact}
The use of generative models in layout design inevitably brings up important questions regarding the ethical use of data. As responsible researchers, we are committed to addressing these concerns with diligence and transparency. During the generation process, we rigorously review the training dataset to ensure full compliance with portrait rights and advertising regulations. Moreover, we engage professional reviewers in the annotation process to meticulously evaluate the generated layouts, ensuring they are free from bias, offensive content, or any violations of applicable laws. Our commitment to ethical standards is central to our research, as we strive to create technology that respects societal norms and values.


    
        

        
        
        


\end{document}